%% file: 00_main.tex
\documentclass[sigconf]{acmart}

\AtBeginDocument{%
  \providecommand\BibTeX{{%
    \normalfont B\kern-0.5em{\scshape i\kern-0.25em b}\kern-0.8em\TeX}}}

\setcopyright{acmcopyright}
\copyrightyear{2023}
\acmYear{2023}
\acmDOI{XXXXXXX.XXXXXXX}

\acmConference[under review]{}{}

\input{macros}
\newcommand{\sysname}{FAIRO\xspace}

\begin{document}

\title[\sysname]{\sysname: Fairness-aware Adaptation in Sequential-Decision Making for Human-in-the-Loop Systems}

\author{Tianyu Zhao}
\email{tzhao15@uci.edu}
\orcid{0009-0000-3139-4872}
\affiliation{%
  \institution{University of California, Irvine}
  \city{Irvine}
  \state{California}
  \country{USA}
}

\author{Mojtaba Taherisadr}
\email{taherisa@uci.edu}
\affiliation{%
  \institution{University of California, Irvine}
  \city{Irvine}
  \state{California}
  \country{USA}
}
\author{Salma Elmalaki}
\email{salma.elmalaki@uci.edu}
\affiliation{%
  \institution{University of California, Irvine}
  \city{Irvine}
  \state{California}
  \country{USA}
}








\renewcommand{\shortauthors}{Zhao \emph{et.al.}}

\begin{abstract}
\input{00_abstract}

\end{abstract}







\settopmatter{printfolios=true}

\settopmatter{printacmref=false} 
\renewcommand\footnotetextcopyrightpermission[1]{} 

\maketitle

\input{01_introduction}
\input{02_relatedwork}

\input{03_motivation}
\input{04_framework}
\input{06_application1Home}

\input{07_application2Water}
\input{08_application3Learning}

\input{09_conclusion}

\begin{acks}
This work is supported by NSF award \#2105084.
\end{acks}

\bibliographystyle{ACM-Reference-Format}
\bibliography{citations}










\end{document}

%% file: macros.tex
\usepackage{xspace}
\usepackage{enumitem}
\usepackage{tabularx}

\usepackage{amsthm}
\usepackage{algorithm}
\usepackage[noend]{algpseudocode}
\usepackage{amsmath}

\usepackage[table]{colortbl}

\usepackage[font=small, skip=0.5ex]{caption}

\renewcommand{\r}{\textcolor{red}} 

\renewcommand{\r}{\textcolor{red}}

\newcommand{\m}{\textcolor{magenta}}

\newcommand\todo[1]{\textcolor{orange}{(TODO: #1)}}

\usepackage{multirow}

%% file: 00_abstract.tex

Achieving fairness in sequential-decision making systems within Human-in-the-Loop (HITL) environments is a critical concern, especially when multiple humans with different behavior and expectations are affected by the same adaptation decisions in the system. This human variability factor adds more complexity since policies deemed fair at one point in time may become discriminatory over time due to variations in human preferences resulting from inter- and intra-human variability. 
This paper addresses the fairness problem from an equity lens, considering human behavior variability, and the changes in human preferences over time. We propose \sysname, a novel algorithm for fairness-aware sequential-decision making in HITL adaptation, which incorporates these notions into the decision-making process. In particular, \sysname decomposes this complex fairness task into adaptive sub-tasks based on individual human preferences through leveraging the Options reinforcement learning framework. We design \sysname to generalize to three types of HITL application setups that have the shared adaptation decision problem. 

Furthermore, we recognize that fairness-aware policies can sometimes conflict with the application's utility. To address this challenge, we provide a fairness-utility tradeoff in \sysname, allowing system designers to balance the objectives of fairness and utility based on specific application requirements. Extensive evaluations of \sysname on the three HITL applications demonstrate its generalizability and effectiveness in promoting fairness while accounting for human variability.
On average, \sysname can improve fairness compared with other methods across all three applications by $35.36\%$.

%% file: 01_introduction.tex
\vspace{-2mm}
\section{Introduction}\label{sec:intro}
The emerging technologies of sensor networks and mobile computing give the promise of monitoring the humans' states and their interactions with the surroundings 
and have made it possible to envision the emergence of human-centered design of Internet-of-Things (IoT) applications in various domains. This tight coupling between human behavior and computing enables a radical change in human life~\cite{picard2000affective}. 
By continuously developing a cognition about the environment and the human state and adapting/controlling the environment accordingly, a new paradigm for IoT systems provides the user with a personalized experience, commonly named Human-in-the-Loop (HITL) systems.  With the increasing number of HITL IoT applications being controlled by artificial
intelligence (AI) algorithms, the algorithmic fairness of such decision-making algorithms has drawn considerable attention in the last few years~\cite{kleinberg2018algorithmic,rambachan2020economic, dwork2012fairness, kusner2017counterfactual}. Nevertheless, the unique nature of HITL IoT opens a new frontier of algorithmic fairness issues that must be carefully addressed before the wide use of such technologies.
In particular, the immense challenge in designing the future HITL IoT lies in respecting human rights and human values, ensuring ethics and fairness, and meeting regulatory guidelines, even while safeguarding our environment and natural resources~\cite{george2023roadmap}.

The following summarizes the key distinctions between the existing literature on algorithmic fairness and the nature of Human-in-the-Loop (HITL) systems:
\begin{itemize}[leftmargin=*, topsep=0pt, noitemsep]
\item \textbf{Fairness in static/singular decision-making vs fairness in dynamic/sequential decision-making:}
The current literature on algorithmic fairness primarily addresses the unfairness arising from biases in data and algorithms used in static systems, often employing supervised learning methods. A canonical example comes from a tool used by courts in the United States to make pretrial detention and release decisions (COMPAS)~\cite{northpointe2015practitioner,angwin2016machine}. Other applications include loan applications~\cite{mukerjee2002multi}, employment processes~\cite{cohen2019efficient}, and markets. In contrast, HITL systems are dynamic, where actions taken at one time have consequences for future states and actions. Therefore, ensuring fairness in HITL systems requires considering the impact of decisions over time, leading to a sequential decision-making problem. Neglecting the dynamic feedback and long-term effects in such systems, as commonly done in static decision-making, can result in harm to sub-populations~\cite{creager2020causal, liu2018delayed, d2020fairness}.

\item \textbf{Fairness in decisions (or equality) vs fairness in the impact of decisions (or equity):}
Existing fairness definitions predominantly focus on equality, aiming to eliminate prejudice or favoritism based on individuals' characteristics. However, insufficient attention has been given to equity, which entails allocating resources to individuals or groups to support their success~\cite{mehrabi2021survey}. Equity becomes crucial in HITL systems. Hence, a shift from fairness defined in terms of equality to fairness based on equity is essential.

\end{itemize}
Motivated by these observations, this paper revisits fairness literature and emphasizes the importance of fairness in sequential decision-making from an equity perspective for HITL systems. The main objective is to operationalize equity in the context of sequential decision-making to develop improved adaptation algorithms tailored to HITL applications.

This paper introduces \textbf{\sysname}, a novel fairness-aware adaptation framework for sequential-decision making designed for HITL systems. The framework specifically tackles the issue of fairness in situations where multiple humans share the same application space and are collectively impacted by adaptation decisions. While usually these decisions aim to optimize overall system performance, they may inadvertently lead to undesired consequences as humans interact with the system or as the system's physical dynamics evolve.


%% file: 02_relatedwork.tex
\vspace{-2mm}
\section{Related Work} \label{sec:rl}
\subsection{Fairness in decision-making systems}
At the heart of HITL systems is achieving the objective of designing scalable, real-time decision-making mechanisms that are aware of the social context, such as the perceived notion of fairness, social welfare, ethics, and social norms~\cite{sztipanovits2019science, khargonekar2020framework}. A vast work in the game theory literature studies various notions of fairness 
between communities by defining incentive markets between competitors to achieve fairness~\cite{ratliff2020adaptive, ratliff2018perspective}. Fairness-enhancing interventions have been introduced to machine learning to ensure non-discriminatory decisions by the trained models~\cite{friedler2019comparative, hashimoto2018fairness,chouldechova2018frontiers, kannan2018smoothed, goel2018non}. In particular, the question of fairness in decision-making systems where the agent prefers one action over another~\cite{jabbari2017fairness, joseph2016fairness, yu2019deep, gillen2018online, siddique2020learning, shin2017exploring} becomes more significant in multi-agent systems~\cite{jiang2019learning, hughes2018inequity}.  
However, imposing fairness constraints as a static, singular decision (as standard supervised learning methods do) while ignoring subsequent dynamic feedback or its long-term effect, especially in sequential decision-making systems, can harm sub-populations~\cite{creager2020causal, liu2018delayed, d2020fairness}. Recent work investigates the long-term effects of Reinforcement Learning (RL). It shows that modeling the instantaneous effect of control decisions for single-step bias prevention does not guarantee fairness in later downstream decision actions~\cite{kannan2019downstream, milli2019social}. Unfortunately, all of this work focused on fairness from the lens of equality---where the target is to ensure no favoritism or bias is present in the system---with very little work that focused on fairness from the lens of equity (mostly in singular/static decision making as opposed to sequential decision-making)~\cite{mehrabi2021survey}. Indeed, achieving fairness in sequential decision-making systems becomes more complex since policies deemed fair at one point may become discriminatory over time due to variations in human preferences resulting from inter- and intra-human factors~\cite{elmalaki2018sentio}. This paper focuses on answering this question, especially for HITL systems.

\vspace{-2mm}
\subsection{Different notions of group fairness}\vspace{-1mm}
The notion of ``group fairness'' is used in the literature to address the fairness problem when multiple humans are affected by the same adaptation model.
While there are several definitions and approaches to defining group fairness, it's important to note that these approaches may have nuanced variations and can be interpreted differently depending on the context and specific application domain. We only summarize two widely used notions of group fairness: (1) equalized odds, which focuses on achieving similar prediction accuracy across different groups while considering binary classification tasks. It ensures that the true positive rate (sensitivity) and true negative rate (specificity) of a predictive model are comparable across different groups~\cite{hardt2016equality}, and (2) equal opportunity: aims to ensure that the predictive model provides an equal chance of benefiting from positive outcomes for all groups. In particular, equal opportunity requires that the true positive rate for each group should be approximately equal~\cite{hardt2016equality}. 
While these two definitions primarily focus on binary classification tasks, this paper will exploit some of their ideas towards sequential decision-making and not specifically for classification tasks. 

\vspace{-2mm}
\subsection{Multi-agent RL and hierarchical RL}\vspace{-1mm}
Reinforcement Learning (RL) has emerged as a widely used approach for monitoring and adapting to human intentions and responses, enabling personalized sequential adaptations in various contexts~\cite{sadigh2017active, hadfield2016cooperative, elmalaki2022maconauto}. To account for individual variability and response times under different autonomous actions approaches like multisample RL and adaptive scaling RL (ADAS-RL) have been proposed~\cite{elmalaki2018sentio, ahadi2021adas}. Amazon has utilized personalized RL to tailor adaptive class schedules based on students' preferences~\cite{bassen2020reinforcement}. Moreover, advancements in deep learning coupled with RL have been leveraged to determine the optimal content to present to students based on their cognitive memory models~\cite{reddy2017accelerating}.
Hierarchical reinforcement learning (HRL) offers a promising solution for addressing complex learning tasks by decomposing them into multiple simpler sub-tasks. This hierarchical structure effectively decomposes long-horizon and intricate tasks into manageable components. The high-level policy is responsible for selecting optimal sub-tasks, considered high-level actions. In contrast, the lower-level policy focuses on solving the sub-tasks using reinforcement learning techniques. This decomposition strategy enables the transformation of a long timescale task into multiple shorter timescale sub-tasks, potentially making individual sub-tasks easier to solve. 
For instance, the Option-critic Framework introduces an architecture capable of learning higher and lower-level policies without needing prior knowledge of sub-goals~\cite{bacon2017option}. HRL has demonstrated superior performance in various domains, including long-horizon games and continuous control problems~\cite{claure2020multi, hoffman2018discretion} 
In this paper, we will decompose the fairness problem into sub-tasks over smaller time horizons and exploit the options framework to solve these sub-tasks.

\vspace{-2mm}
\subsection{Paper contribution} 
The contributions of this paper can be summarized as follows:
\begin{itemize}[noitemsep, topsep=0pt, leftmargin=*]
\item \textbf{Fairness from the Lens of Equity:} We tackle the fairness problem in sequential-decision making systems within HITL environments by addressing the notion of equity. 
\item \textbf{\sysname:} We propose \sysname, a novel algorithm designed for fairness-aware sequential-decision making in HITL adaptation. Our approach leverages the Options RL framework to effectively incorporate fairness considerations into the decision-making process.
\item \textbf{Generalization to Different HITL Application Setups:} We extend \sysname to cater to three types of HITL application setups. These setups involve multiple humans sharing the application space and being impacted by: (1) global numerical adaptation decisions, (2) shared global resources, and (3) shared global categorical adaptation decisions.
\item \textbf{Evaluation on Multiple HITL Applications:} We conduct comprehensive evaluations of \sysname on three different HITL applications to demonstrate its generalizability.
\end{itemize}

The paper is structured as follows: Section~\ref{sec:summaryoptionsframework} summarizes the Options framework, which serves as the foundation for our proposed approach. 
Section~\ref{sec:frameworkmodel} details how we leverage the Options framework to incorporate fairness considerations into the decision-making process. The subsequent sections of the paper focus on evaluating our proposed approach, \sysname, in three distinct application domains. 


%% file: 04_framework.tex
\vspace{-2mm}
\section{Options framework for temporal abstraction}\label{sec:summaryoptionsframework}

The Markov Decision Process (MDP) framework is widely employed for modeling sequential decision-making. Various methods are utilized to solve MDPs and obtain the optimal Markov decision chain, including dynamic programming and reinforcement learning (RL). RL is particularly used when the transition probabilities within the MDP are unknown. Within the discrete-time finite MDP setting, the standard RL framework can be applied. In particular, an \emph{agent} engages with an \emph{environment} that is modeled as an MDP at discrete time steps, denoted as $t = 0, 1, 2, ...$. At each time step $t$, the agent observes the current state of the environment, denoted as $s_t \in \mathcal{S}$, and selects an action $a_t \in \mathcal{A}$ based on this observation. This action leads to a transition to the next state, $s{t+1}$, and yields a reward value, $r_t \leftarrow \mathbb{R}$, associated with this transition. By engaging in this interaction, the agent learns a policy $\pi(s,a)$ that guides its decision-making, aiming to select the best action $a$ for each state $s$ to maximize the expected total reward over sequential decision actions.

\begin{figure}[!t]
\centering
\includegraphics[width=1\columnwidth]{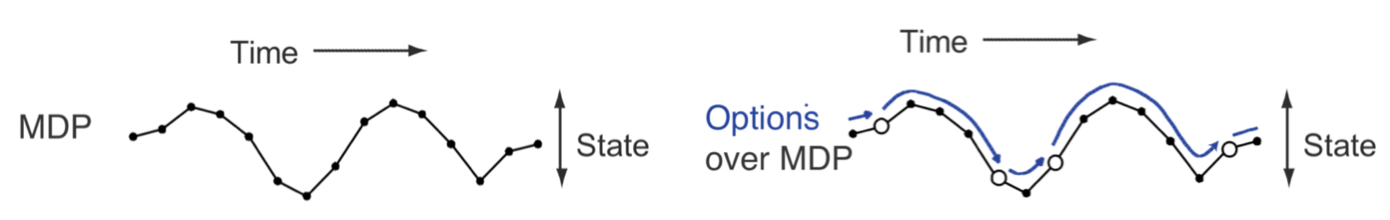}
\caption{The state trajectory of an MDP with small discrete-time transitions. Options enable overlaid larger abstracted discrete events.}
\label{fig:optMDP}\vspace{-5mm}
\end{figure}

The options framework was first introduced by Sutton et al. ~\cite{sutton1999between} to generalize primitive actions to include temporally extended courses of lower-level action. In particular, the term \text{options} represents a temporal abstraction of the lower-level actions in the MDP. A pictorial figure of options over MDP is shown in Figure~\ref{fig:optMDP}. An MDP's state trajectory comprises small, discrete-time transitions, whereas the options enable an MDP to be abstracted and analyzed in larger temporal transitions. 

Option $o$ within the option set $\mathcal{O}$ consists of three main components: a policy $\pi(a|s, o)$ for selecting actions within option $o$, an initiation set $\mathcal{I} \subseteq \mathcal{S}$, a termination condition $\beta$. 
An option $o$:($\mathcal{I}, \pi, \beta$) is available to be selected by the agent in state $s_t$ if and only if $s_t \in \mathcal{I}$. If the option is selected, actions are selected according to the option policy $\pi$ until the option terminates according to the termination condition $\beta$. When the option terminates, the agent can select another option. This definition of options makes them act as much like actions while adding the possibility that they are temporally extended\footnote{Options framework can be extended to include policies over options. In particular, when multiple options are available to the agent at $s_t$, the agent can learn which option to select using the policy over options. In this paper, we consider the policy over options to be a fixed policy, and the initiation sets of all options are disjoint sets. }.


The rationale behind employing the options framework to achieve fairness in a multihuman setting stems from the inherent limitations imposed by an option's initiation set $\mathcal{I}$ and termination condition $\beta$. These constraints confine the applicability of an option's policy, $\pi$, to a subset defined by $\mathcal{I}$ rather than encompassing the entire state space $\mathcal{S}$. Consequently, options can be viewed as a means of achieving \textbf{fairness subgoals}, wherein each option's policy is adapted to enhance the attainment of its specific \textbf{subgoal}, thereby contributing to the overall fairness of the decision-making agent. Notably, the dynamic nature of the multihuman environment necessitates the need for diverse fairness policies at different temporal instances. Consequently, in this paper, we leverage the options framework to address fairness concerns within the context of sequential decision-making.

\begin{figure*}[!t]
\centering
\includegraphics[width=1\linewidth]{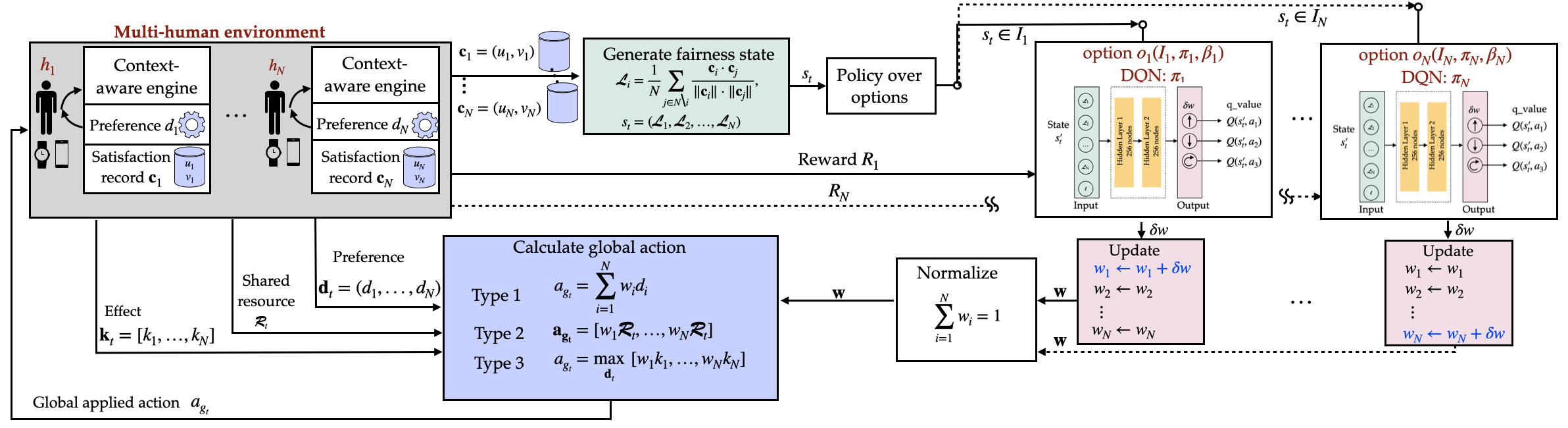}
\caption{\sysname for fairness-aware framework in Human-in-the-Loop systems using options framework. \sysname is designed for three types of applications: Type 1: one global action based on numerical demands affecting multiple humans, Type 2: one shared resource distributed over multiple humans, and Type 3: one global action based on categorical preferences. }
\label{fig:framework}\vspace{-5mm}
\end{figure*}

\vspace{-1mm}
\section{Fairness using Options framework}\label{sec:frameworkmodel} 
We exploit the options framework to design \sysname to achieve fairness in sequential decision-making agents in multihuman environment. 
As seen in Figure~\ref{fig:framework}, the agent interacts in sequential discrete-time steps with an environment that has $N$ humans $(h_1, h_2,\dots, h_N)$ through observing their preferences or their desired adaptation actions $(d_{1}, d_{2}, \dots, d_{N})$ and the current fairness state of the environment $s_t$. Guided by the current fairness state $s_t$, the agent selects an appropriate option $o_t$ from the set of $N$ available options $\mathcal{O}$. The chosen option $o_t$ then determines a lower-level action based on its specific option policy $\pi_o$, resulting in a global action $a_{g_t}$ that is applied to the shared environment. This global action subsequently modifies the current fairness state, and the agent receives a reward $r_{t+1}$. This reward is utilized to refine the option policy. In the following subsections, we provide a detailed description of each module within the \sysname framework. 

\vspace{-2mm}
\subsection{Fairness state space $\mathcal{S}$}
Our approach to viewing fairness from the lens of equity is by using a fairness state that encompasses the history of the positive and negative effects of the global decision action.   

\vspace{-2mm}
\subsubsection{\textbf{Satisfaction history records $\mathbf{c}_i$} }
Fairness state $s_t$ is inferred from the history of the \emph{satisfaction} of each human. To model the satisfaction of the human $h_i$, we keep a history record for each human:
\begin{equation}
\mathbf{c}_i = ( u_i, v_i ),  \text{where } i \in \{1, 2, \dots, N\} . 
\end{equation}
The value $u_i \in \mathbb{R}$ represents a record of the number of times the human $h_i$ was unsatisfied by the applied global action $a_{g}$. In contrast, $v_i \in \mathbb{R}$ represents a record for the number of times the human $h_i$ was satisfied by the applied global action $a_{g}$.

At time step $t$, every human $h_i$ has a desired adaptation action $d_{i_t}$. For example, a human may prefer a particular temperature setpoint to HVAC system (Heating, ventilation, and air conditioning) in their room for thermal comfort that matches their physical activities, such as sleeping, domestic work, or sitting~\cite{elmalaki2018internet}. Based on the difference in the values of $d_{i_t}$ and $a_{g_t}$, the record $\mathbf{c}_i$ is updated to capture whether the human was satisfied or unsatisfied. For example, if this difference is within a threshold $\tau$ then we consider the human $h_i$ is satisfied and increment $v_i$ by a value $\delta$. 
\begin{equation}\label{eq:updatecounter}
\mathbf{c}_{i} = 
\begin{cases}
(u_i, v_i+ \delta) & \|d_{i_t} - a_{g_t}\| \leq \tau. \\
(u_i+ \delta, v_i) & \|d_{i_t} - a_{g_t}\| >  \tau.
\end{cases}
\end{equation}

After all the records $\mathbf{C} = (\mathbf{c}_i , i = 1, 2, ..., N)$ are updated, they are normalized to a unit vector. Choosing the value $\tau$ is application dependent; however, the value $\delta$ needs to be less than $1$ and small enough to ensure that the unit vector direction $\mathbf{C}$ does not change drastically. Hence, we choose $\delta$ to be $0.01$. 

Ideally, these records $\mathbf{c}_i$ should be $( 0 , 1 )$ indicating that the global adaptation action $a_{g}$ meets the preferences of the human over time. However, as we mentioned earlier, these preferences may conflict with humans sharing the same environment. Hence, the same $a_{g_t}$ may be perceived by one human as meeting their preference (increasing $v$) and by another human as not meeting theirs (increasing $u$).

\begin{figure}[!t]
\centering
\includegraphics[width=0.5\columnwidth]{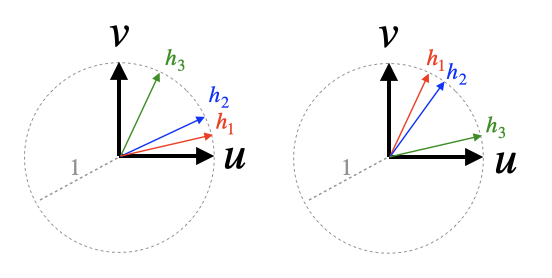}
\caption{Satisfaction history record $\mathbf{C}$ is represented as $2D$ unit vector, where the two components $v$ and $u$ represent the value that correlates to the number of time steps the human received an adaptation action to the shared environment $a_g$ that is close to the preferred desired action $d$ within a threshold $\tau$ or far from it respectively. }\vspace{-2mm}
\label{fig:counters}\vspace{-4mm}
\end{figure}

A pictorial visualization of $\mathbf{C}$ is shown in Figure~\ref{fig:counters}. Each $\mathbf{c}_i$ can be represented as a $2D$ vector within a unit circle. We show two examples for $\mathbf{c}_i$ for three humans where $h_3$ has $\mathbf{c}_3$ closer to the $v$ axis compared to the other two humans (Figure~\ref{fig:counters}-left) versus the case when $h_3$ has $\mathbf{c}_3$ closer to the $u$ compared to the other two humans (Figure~\ref{fig:counters}-right).  
Figure~\ref{fig:counters} shows an example of a relatively unfair situation, where $h_3$ is either treated most of the time favorably (Figure~\ref{fig:counters}-left) or unfavorably (Figure~\ref{fig:counters}-right).

It is worth mentioning here that $\mathbf{C}$ captures the history of the effect of the trajectory of sequential adaptation action on the shared environment. 
Hence, the intuition is to tune the global action in the next time step $a_{g_{t+1}}$ to either decrease the $focus$ on considering the preferences of $h_3$ (Figure~\ref{fig:counters}-left) or vice versa (Figure~\ref{fig:counters}-right). However, as mentioned in Section~\ref{sec:intro}, the same action affects all the humans sharing the same environment.
\vspace{-2mm}
\subsubsection{\textbf{Fairness state $s_t$}}\label{sec:fairnessstate} 
We use the geometric intuition in Figure~\ref{fig:counters} to design our fairness state $s_t$ to compare the directions of all $N$ records in $\mathbf{C}$. Ideally, we would like to have all $\mathbf{c}_i$ as close as possible to each other. Hence, we define $s_t$ to capture how close each $\mathbf{c}_i$ is to the other $N\text{\textbackslash} i$ records. Hence, we define ($s_t$) as follows:  
\setlength{\abovedisplayskip}{3pt}
\setlength{\belowdisplayskip}{3pt}
\begin{align}
\mathcal{L}_i &= \frac{1}{N} \sum_{j \in N \text{\textbackslash} i}  \frac{\mathbf{c}_i \cdot \mathbf{c}_j} {\| \mathbf{c}_i\| \cdot \|\mathbf{c}_j \|}, & \mathcal{L}_i \in ]0,1] \subset \mathbb{R}\\
s_t &= ( \mathcal{L}_1, \mathcal{L}_2, \dots, \mathcal{L}_N ), & s_t \in \mathcal{S}=]0,1]^N 
\label{eq:state}
\end{align}

In particular, $\mathcal{L}_i$ represents the closeness of record $\mathbf{c}_i$ to the rest of the records using the average of the cosine of the angle between pair of vectors. 
Hence, if the cosine value between two vectors is $1$, they coincide. 
Since the values of $\mathbf{c}_i$ can only be positive and are normalized to a unit vector, the minimum cosine value between these vectors is $0$, indicating that they are far from each other (at $90^\circ$)\footnote{$\mathbf{c}_i \in \mathbb{R}, \mathcal{L}_i$ is unlikely to reach $0$ but can decrease to a very small value $\epsilon$. 
}. Equation~\ref{eq:state} represents $s_t$ which holds all the values of $\mathcal{L}_i$. Ideally, from our fairness point of view, the goal state $s_t$ should be $( 1, 1, \dots 1 )$, which indicates that all $\mathbf{C}_h$ have the same direction, meaning that the history of the satisfaction and unsatisfaction for all the humans are close.

\vspace{-3mm}
\subsection{Initiation set $\mathcal{I}$ and fairness subgoals}\vspace{-1mm}
While the ultimate goal is to learn a policy that can achieve the goal state $s_t = ( 1, 1, \dots 1 )$, this is challenging since it is a huge state space. Accordingly, the intuition behind exploiting the options framework is to divide this goal into smaller subgoals where we learn over a subset of states or the initiation set ($\mathcal{I} \subseteq \mathcal{S}$) as explained in Section~\ref{sec:summaryoptionsframework}. We divide $\mathcal{S}$ into $N$ initiation sets $\mathcal{I}_i$ where $i \in \{1,2,\dots,N\} $, such that $\mathcal{I}_i$ contains all the states with $\mathcal{L}_i$ as the minimum value.
\setlength{\abovedisplayskip}{3pt}
\setlength{\belowdisplayskip}{3pt}
\begin{equation}
\mathcal{I}_i = \{ s_t = \{ \mathcal{L}_1, \ldots, \mathcal{L}_N \} \in \mathcal {S}| \mathcal{L}_i = \min (s_t)   \}  
\end{equation}

Specifically, this means that each initiation set $\mathcal{I}_i$ considers only the states where $h_i$ has received unfair adaptation either favorably or unfavorably. For example, both cases in Figure~\ref{fig:counters} are considered unfair state where $\mathcal{L}_3$ is less than $\mathcal{L}_1$ and $\mathcal{L}_2$. 

\vspace{-3mm}
\subsection {Termination State $\beta$}
Each option $o_i$ terminates when the current state $s_t$ reaches a terminal state for this option. Hence, in \sysname, the set of terminal states for $o_i$ is when $\mathcal{L}_i$ is no longer the minimum value in $s_t$.
\setlength{\abovedisplayskip}{3pt}
\setlength{\belowdisplayskip}{3pt}
\begin{equation}
\beta_i = \{ s_t = \mathcal{L}_1, \ldots, \mathcal{L}_N \} \in \mathcal {S}| \mathcal{L}_i \neq \min (s_t) \}
\end{equation}

Intuitively, this means that each option $o_i$ will run to improve the value of $\mathcal{L}_i$ until it is no longer the minimum value which is the fairness subgoal for this option. This will trigger a new initiation set $I$ and this option terminates and a new option starts to achieve another subgoal: improving $\mathcal{L}_i$.


\vspace{-2mm}
\subsection{Global action of different HITL applications}\label{sec:apptypes}

As shown in Figure~\ref{fig:framework}, every human ($h_i$) has a desired preference ($d_i$) for the adaptation action. However, only one action $a_g$ is chosen to be applied to the shared environment. In \sysname, to learn the global action $a_g$ that can achieve a better fairness state $s_t$ that eventually achieves the fairness subgoal, we categorize the types of applications into three types: 
\begin{itemize}[noitemsep, topsep=0pt, leftmargin=*]
\item \textbf{(Type 1) Shared numerical global action:} The desired preferences $d_i$ have numerical values and the global action $a_g$ is a numerical value. In this case, we design each option to take a weighted sum of these $N$ preferences. These different weights represent the contribution of each human preference $d_i$ in the applied global action $a_{g_t}$. We will show an instant of this type in a simulated smart home application in Section~\ref{sec:evalhome}.
\begin{equation}\label{eq:globalaction}
\begin{split}
a_{g_t} &= \sum_{i=1}^N w_i  d_i, \ \ \ \  w_i \in [0,1]
\end{split}
\end{equation}

\item \textbf{(Type 2) Shared global resource:} The desired preferences $d_i$ have numerical values. There is one resource $\mathcal{R}$ that is shared, time-varying, and it has to be distributed. The global action $\mathbf{a_g}_t$ is the weighted share of this resource $\mathcal{R}_t$ dictated by their desired preferences. We will show an instant of this type in a simulated water distribution application in Section~\ref{sec:evalwater}.
\begin{equation}\label{eq:globalresource}
\begin{split}
\mathbf{a_g}_t &= [w_1  \mathcal{R}_t, w_2 \mathcal{R}_t, \dots, w_N  \mathcal{R}_t ], \ \ \ \  w_i \in [0,1]
\end{split}
\end{equation}

\item \textbf{(Type 3) Shared categorical global action:} The desired preferences $d_i$ have categorical values and the global action $a_{g}$ is one of these categorical values. 
In this case, the global action selects one of the $d_i$ with the maximum  weighted effect. In this work, we define the effect of $d_i$ as $k_i$ to quantify applying $d_i$ on all the other $N-i$ humans. This effect can be estimated from the context-aware engine as shown in Figure~\ref{fig:framework}.  We will show an instant of this type for smart education application in Section~\ref{sec:evallearning}.
\setlength{\abovedisplayskip}{3pt}
\setlength{\belowdisplayskip}{3pt}
\begin{equation}\label{eq:globalcategory}
\begin{split}
\mathbf{a_g}_t &= \max_{d_i}[w_1 k_1, w_2 k_2, \dots, w_N  k_N ], \ \ \ \  w_i \in [0,1]
\end{split}
\end{equation}
\end{itemize}

\vspace{-3mm}
\subsection{Learning the option policy ($\pi_i$)} 
Each option policy $\pi_o$ learns the appropriate $\mathbf{w} = (w_1, w_2, \dots, w_N)$ for the global action $a_{g_t}$ for every state $s_t \in \mathcal{I}_i$ to reach the termination condition $\beta_i$ such that $\sum_{i=1} ^N w_i = 1$.  These weights are continuous values and learning them for every $s_t \in \mathcal{I}_i$ is challenging. Hence, we opt for a simpler design of using Deep Q-Network (DQN) to reduce the search space for the appropriate weights as detailed below.
\vspace{-2mm}
\subsubsection{\textbf{Deep Q-Network (DQN)}} \label{sec:dqnintro}
\begin{figure}[!t]
\centering
\includegraphics[width=0.7\columnwidth]{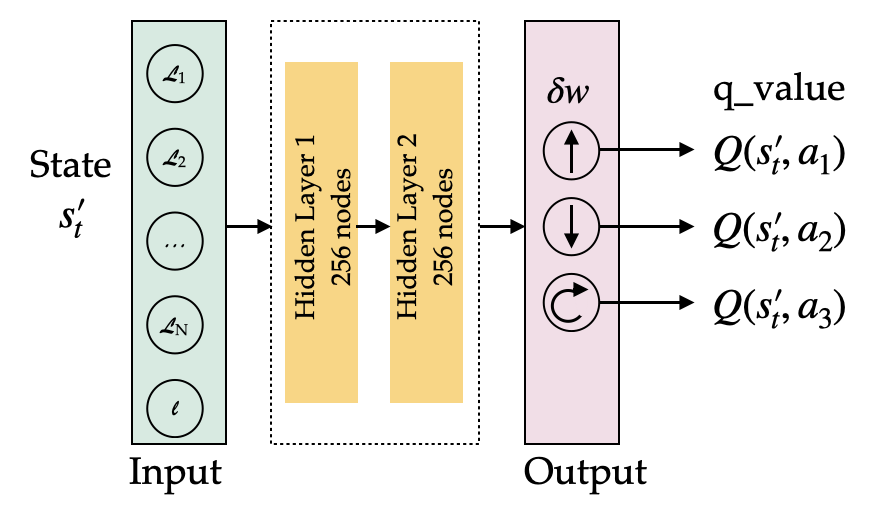}
\caption{DQN for option $o_i$. The input is the state $s_t'$. The output is the \texttt{q\_value} for the possible three actions of weight adjustment $\delta w$.}
\label{fig:DQN}
\end{figure}

DQN is a reinforcement learning algorithm that combines Q-learning with deep neural network~\cite{mnih2013playing}. In particular, DQN is used with Markov Decision Process (MDP) to learn the best action to apply at a particular state.
DQN estimates the action-value function, denoted as $Q(s, a)$, which represents the expected return (reward) for taking action $a$ in state $s$. 
This is typically represented by a neural network, where the inputs are the state $s$ and the outputs are the estimated action values $Q(s, a)$ for each possible action. At each time step $t$, the DQN agent observes the state $s_t$ and selects an action $a_t$ based on its current estimate of the Q-function $Q(s, a)$. This can be done through an $\varepsilon$-greedy policy, where the action with the maximum estimated value is selected with a probability of $1 - \varepsilon$ and a random action is selected with probability $\varepsilon$. 
The DQN agent then applies the action, observes the next state $s_{t+1}$ and the reward $r_{t+1}$, and updates its estimate of the Q-function using the following update rule~\cite{sutton2018reinforcement}:
\setlength{\abovedisplayskip}{3pt}
\setlength{\belowdisplayskip}{3pt}
\begin{equation}    
    \begin{split}
        Q(s_t, a_t) \leftarrow & Q(s_t, a_t) + \alpha(r_{t+1} +  \gamma \max_a Q(s_{t+1}, a) - Q(s_t, a_t)),
    \end{split}
\end{equation}

\noindent where $\alpha$ is the learning rate, $\gamma$ is the discount factor, and $\max_a Q(s_{t+1}, a)$ is the estimated maximum action-value in the next state $s_{t+1}$.
DQN aims to learn a policy that maximizes the cumulative reward over time in a given environment. It combines Q-learning with deep learning, allowing the agent to handle high-dimensional observations and non-linear function approximations. Figure~\ref{fig:DQN} shows a pictorial illustration for the design of the DQN in every option $o_i$ in \sysname.

\vspace{-2mm}
\subsubsection{\textbf{Input to Option DQN ($s_t'$)}}
Every option $o_i$ runs a DQN where the input is the $s_t$. However, to differentiate between the two cases shown in Figure~\ref{fig:counters} where both $\mathcal{L}_3$ is the minimum value in $s_t$, we append to $s_t$ the relative location of $\mathbf{c}_3$ with respect to the $\mathbf{c}_1$ and $\mathbf{c}_2$.

\vspace{-5mm}
\begin{align}\label{eq:optionstate}
s_t' = (s_t, l), \text{ where } s_t \in \mathcal{I}_i, \quad
l = 
\begin{cases}
 1,&  v_i = \max_{v} (s_t).\\ 
 0,& \text{otherwise}.
\end{cases}
\end{align}

This means that if $\mathcal{L}_i$ is the minimum in $s_t$, option $o_i$ will run since $s_t \in \mathcal{I}_i$. The input to the DQN in option $o_i$ will indicate whether $\mathcal{L}_i$ has the minimum value (unfair situation) because $h_i$ received favorable treatment relative to all $h_{N\textbackslash i}$ (\emph{i.e., }$\mathbf{c}_i$ has a high $v_i$ component) and in this case $l=1$ in Equation~\ref{eq:optionstate}, or $l=0$ otherwise. 

\vspace{-2mm}
\subsubsection{\textbf{Output from Option DQN ($\delta w_i)$}} 
To reduce the action space of the DQN since we need to learn $N$ weights with continuous values $[0,1]$, we designed the output from the DQN in option $o_i$ to focus only on adjusting $w_i$ instead of all the $N$ weights $\mathbf{w}$. In particular, as shown in Figure~\ref{fig:DQN}, the output from the DQN are the Q-values ($Q(s,a)$) which decides to select between three actions to ($a_1$) increase the weight $w_i$ $\uparrow$, ($a_2$) decrease the weight $w_i$ $\downarrow$, or ($a_3$) keep the same weight $w_i$ $\circlearrowright$. Hence, the output from DQN (which is the selected action of the DQN as explained in Section~\ref{sec:dqnintro}) is the weight adjustment for $w_i$ by $\delta w_i = (+\delta, -\delta, 0 )$ where the value $\delta$ determines how fast the DQN for option $o_i$ changes the weight $w_i$. 
\begin{align}\label{eq:weightadj}
w_i \leftarrow w_i + \delta w_i
\end{align}

Since all the weights $\mathbf{w}$ have to be normalized to $1$, all the weights will be adjusted accordingly. Using this design, we choose between $3$ possible adjustments for weight $w_i$ instead of the whole space $[0,1]$ and instead of all the $N$ weights. Accordingly, at every time step $t$, each weight in $\mathbf{w}$ is adjusted due to the normalization. 
\setlength{\abovedisplayskip}{2pt}
\setlength{\belowdisplayskip}{3pt}
\begin{equation}
\sum_{i=1}^N w_i = 1, \qquad
\mathbf{w} \leftarrow  \mathbf{w} + \Delta \mathbf{w}
\end{equation} \vspace{-2mm}




\vspace{-2mm}
\subsubsection{\textbf{Option DQN reward ($R_i$)}}
Each option DQN learns the appropriate policy $\pi_i (s_t', \delta w_i)$, which is the right weight adjustment $\delta w_i$ at state $s_t^i$. The DQN learns this policy through a notion of a feedback reward as explained in Section~\ref{sec:dqnintro}. As each option $o_i$ aims at increasing the fairness subgoal of enhancing $\mathcal{L}_i$ while improving the performance of the application, the reward function $R_i$ can be expressed with two terms; a fairness term $\mathcal{F}$, and a performance term $\mathcal{P}$ using a trade-off parameter $\zeta \in[0,1]$. 
\setlength{\abovedisplayskip}{3pt}
\setlength{\belowdisplayskip}{3pt}
\begin{equation}\label{eq:rewardoption}
\begin{split}
 R_{i} &= \zeta \mathcal{F}_i + (1-\zeta) \mathcal{P}_i,  \in [-1, 1] \subset \mathbb{R}\\
 \mathcal{F}_i &= \text{absolute fairness} +  \text{option improvement} \\ & = (2*\mathcal{L}_{i_t} - 1) + f(\mathcal{L}_{i_{t-1}},\mathcal{L}_{i_{t}} ) ,\in [-1, 1] \subset \mathbb{R} \\
 \mathcal{P}_i  &= \text{application dependent}, \in [-1, 1] \subset \mathbb{R} 
 \end{split}
 \end{equation}

In particular, the fairness $\mathcal{F}_i$ considers the current value of $\mathcal{L}_i$, which we call the ``absolute fairness'', and the ``improvement in the fairness'' value of $\mathcal{L}_i$ from last time step for this particular option. It is a function ($f$) of the current value of $\mathcal{L}_{i_t}$ and the value from last time step $\mathcal{L}_{i_{t-1}}$ as shown in Equation~\ref{eq:rewardoption}.

The overall \sysname algorithm is listed in Algorithm~\ref{alg:fairoalg}.

\setlength{\textfloatsep}{0pt}
\begin{algorithm}[!t] \small
\caption{\sysname algorithm}\label{alg:fairoalg}
 \begin{algorithmic}[1]
 \Require 
 \Statex Humans $\mathcal{H} = (h_1, \dots, h_N)$ 
 \Statex Satisfaction records $\mathbf{C} =  (\mathbf{c_1}, \dots, \mathbf{c_N}), 
 \mathbf{c}_i = (u,v) \in \mathbb{R}^2$
\Statex States $ s \in \mathcal{S} = ] 0, 1] ^N \subset \mathbb{R}^N$
\Statex Initiation sets $\mathcal{I}_i \in \mathcal{I} = \{s \in \mathcal{S} \}$
\Statex Options $\mathcal{O} = (o_1, \dots, o_N), o_i = (\mathcal{I}_i, \pi_i, \beta_i)$
\Statex Application type $T=\{ Type 1, Type 2, Type 3\}$
 
\Procedure{Run-FAIRO}{}
\While{True}
\State $\mathbf{d}_t \gets$ context-aware-engine ($\mathcal{H}$)
\If{$T == Type 2$} 
    \State $\mathcal{R}_t \gets$ get-current-available-resource()
\EndIf 
\If{$T ==Type 3$}
    \State $\mathbf{k}_t \gets$ get-current-effects($\mathbf{d}_t$)
\EndIf

\State $s_t \gets$ get-fairness-state($\mathbf{C}$)
\State $s_t' \gets$ append-state($s_t$)  
\State $o_t \gets$ choose-option ($s_t$) \Comment $o_t \in \mathcal{O}$
\State $\delta w \gets$ run-option($o_t$)  \Comment Get the weight adjustments \\ \hfill \Comment based on option policy $\pi_o(s_t, \delta w)$ 
\State $\mathbf{w}_t \gets$ update-normalize-weights($\delta w$)
\State $a_{g_t} \gets$ calculate-global-action($\mathbf{d}_t$, $\mathbf{w}_t$,$\mathcal{R}_t$, $\mathbf{k}_t$ )
\State \Comment Apply global action $a_{g_t}$ on the shared environment 
\State $R_t \gets$ receive-reward ()
\State $\pi_o(s_t, \delta w) \gets$ Update-option-policy ($R_t$)
\State $\mathbf{C} \gets $ Update satisfaction records ($\mathbf{d}_t, a_{g_t}$)
\EndWhile
\EndProcedure
 \end{algorithmic}
\end{algorithm}

%% file: 06_application1Home.tex
\vspace{-2mm}
\section{Application Type 1: Smart Home HVAC}\label{sec:evalhome}\vspace{-1mm}
Recent literature focuses on enhancing human satisfaction in smart heating, ventilation, and air conditioning (HVAC) systems by employing reinforcement learning (RL) techniques to adjust the set-point based on human activity and preferences~\cite{jung2017towards,elmalaki2021fair}. These HITL systems consider the current state and individual preferences, such as body temperature changes during sleep or physical activity. 
To evaluate \sysname in Type 1 applications, we consider a setup where multiple humans share a house with a single HVAC system, and their activities determine individual set-point preferences.




\begin{table*}[ht]
    \centering
\begin{tabularx}{\linewidth}{XXX}
\includegraphics[width=\linewidth]{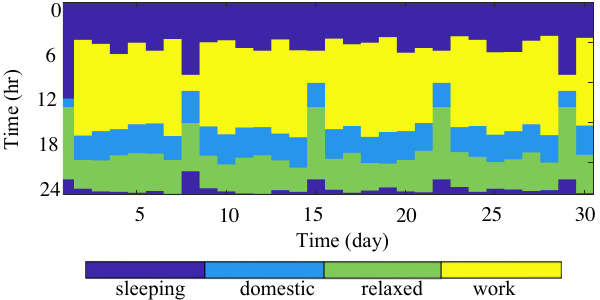}
    \captionof{figure}{Human 1 activity pattern.}\label{fig:act1}
& \includegraphics[width=\linewidth]{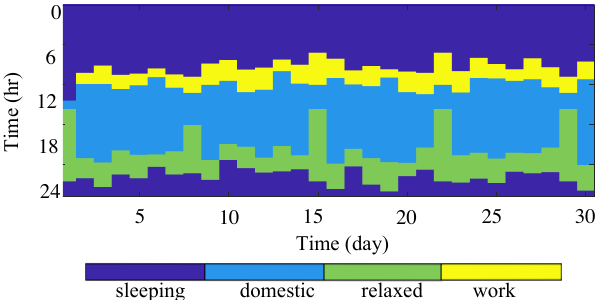}   
    \captionof{figure}{Human 2 activity pattern.}\label{fig:act2}    
& \includegraphics[width=\linewidth]{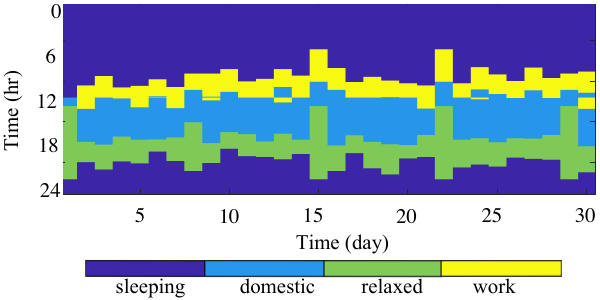}
     \captionof{figure}{Human 3 activity pattern.}\label{fig:act3} 
\end{tabularx}
\vspace{-7mm}
\label{tab:humanactivityapp1}
\end{table*}



\vspace{-3mm}
\subsection{House-Human Model} \label{sec:househumanmodel}

We used a thermodynamic model of a house incorporating the house's shape 
and insulation type. 
To regulate indoor temperature, a heater and a cooler with specific flow temperatures ($50^{\circ}c$ and $10^{\circ}c$) were employed. A thermostat maintained the indoor temperature within $2.5^{\circ}c$ around the desired set point. An external controller controls the setpoint. The human was modeled as a heat source, with heat flow dependent on the average exhale breath temperature ($EBT$) and the respiratory minute volume ($RMV$). 
These parameters depend on human activity~\cite{carrollpulmonary2007}. We simulated three humans with four activities: sleeping, relaxing, medium domestic work, and working from home. 
The different activity schedules depicted in Figures \ref{fig:act1}, \ref{fig:act2}, and \ref{fig:act3}.  
The humans were simulated in separate rooms, each exhibiting unique behavioral patterns: (1) $h_1$ followed an organized and repetitive weekly routine, (2) $h_3$ had a more random and unpredictable life pattern, and (3) $h_2$ displayed intermediate randomness, alternating between sleeping, being away from home, domestic activities, and relaxation.  The Mathworks thermal house model was extended to include a cooling system and a human model\footnote{While more complex simulators like EnergyPlus~\cite{gerber2014energyplus} exist, considering energy consumption and electric loads, we opted for a simpler model to assess \sysname.}.


\vspace{-2mm}
\subsection{Context-aware engine} \label{sec:app1RLdesign}
In Figure~\ref{fig:framework}, a context-aware engine estimates the desired action $d_i$ per human $h_i$ in a smart home. The desired action can be obtained through fixed policy configuration or learned policy. To focus on our main contribution and not on designing a new context-aware engine, we leverage existing RL-based approaches~\cite{taherisadr2023adaparl} for estimating the desired HVAC setpoint $d_i$ based on activity and thermal comfort. The desired setpoints for the considered activities are domestic activity ($72^\circ$F), relaxed activity ($77^\circ$F), sleeping ($62^\circ$F), and work from home ($67^\circ$F). These setpoints aim to enhance thermal comfort. Thermal comfort is assessed using Prediction Mean Vote (PMV) on a scale from very cold ($-3$) to very hot ($+3$)\cite{fanger1970thermal}. Optimal indoor thermal comfort falls within the recommended range of $[-0.5, 0.5]$, as per the ISO standard ASHRAE $55$~\cite{handbook2009american}.

\vspace{-2mm}
\subsection{Evaluation} \label{sec:Evaluationapp1}

We compare $5$ different approaches to determine the HVAC setpoint: 
\begin{itemize}[leftmargin=*]
\item \textbf{\sysname:} 
The global applied action is $a_{g_t} = \sum_{i=1}^3 w_i d_i$ (Equation~\ref{eq:globalaction}).
The reward per option $i$ is as explained in Equation~\ref{eq:rewardoption}. We set $\zeta = 0.5$. As for the performance term $\mathcal{P}_i$ in the reward, we assign a high reward when the $PMV$ falls in the acceptable range $[-0.5,0.5]$. Further, we used the values of the satisfaction counters $\mathbf{c}_i=(u_i,v_i)$ as an indication of the performance $\mathcal{P}_i$ since their values are correlated to the desired temperature $d_i$ which maps to the best PMV. 
The value $\mathcal{P}_i$ is then normalized to be $[-1,1]$: 
\setlength{\abovedisplayskip}{3pt}
\setlength{\belowdisplayskip}{3pt}
\begin{equation}\label{eq:rewardoptionapp1}
\begin{split}
 f&(\mathcal{L}_{i_{t-1}},\mathcal{L}_{i_t}) = sign (\mathcal{L}_{i_{t-1}}-\mathcal{L}_{i_t}) \times Z \\
 Z&= \begin{cases} 
 0 & |\mathcal{L}_{i_{t-1}}-\mathcal{L}_{i_t}| \in ]0,0.001]\\
 0.25 & |\mathcal{L}_{i_{t-1}}-\mathcal{L}_{i_t}| \in ]0.001,0.005]\\
 0.5 & |\mathcal{L}_{i_{t-1}}-\mathcal{L}_{i_t}| \in ]0.005,0.01]\\
 0.75 & |{L}_{i_{t-1}}-\mathcal{L}_{i_t}| \in ]0.01,0.015] \\
 1 & |{L}_{i_{t-1}}-\mathcal{L}_{i_t}| > 0.015 \\
 \end{cases}\\
 \mathcal{P}_i  &=  0.2\frac{v_i}{u_i+v_i} + 0.8f(PMV), \in [-1, 1] \subset \mathbb{R} \\
\end{split}
\end{equation}

\item \textbf{Average approach:} The setpoint is the mean value of desired setpoints of all rooms. Hence, $a_{g_t} = \frac{1}{3}\sum_{i=1}^3 d_i$.

\item \textbf{Equality using Round Robin (RR):} The setpoint is selected from one of the desired setpoints of all rooms in a rotation. The intuition of this approach is to check the case where we give every room the same opportunity to use its desired setpoint across time for equality and how this will affect the overall fairness goal. Hence, for 3 humans, $N=3$, and $a_{g_1}=d_1, a_{g_2}=d_2, a_{g_3}=d_3, a_{g_4}=d_1,  a_{g_5}=d_2\dots, $ etc.

\item \textbf{No subgoals using $1$ DQN with $4$ inputs:} The setpoint is calculated using a single model of 1 DQN with the same structure as Figure~\ref{fig:DQN}. The inputs are the fairness state ($s_t = (\mathcal{L}_1, \mathcal{L}_2, \mathcal{L}_3$) 
plus one input which is an indicator of the room number with the minimal value of $\mathcal{L}$. The intuition of this approach is to evaluate whether we can achieve the overall fairness goal without the need for subgoals that the options framework provides. In this case, the reward for the single DQN is the average reward value across all rooms: %
\setlength{\abovedisplayskip}{3pt}
\setlength{\belowdisplayskip}{3pt}
\begin{equation}\label{eq:reward1dqn}
\begin{split}
\mathcal{R} &= 0.5 \mathcal{F} + 0.5 \mathcal{P},\ \ \ \  \text{where } N = 3 \in [-1, 1] \subset \mathbb{R}\\
\mathcal{F} &= \frac{1}{N}\sum_{i=1}^N \mathcal{L}_i^N + f (\frac{1}{N}\sum_i \mathcal{L}_{i_{t-1}}, \frac{1}{N}\sum_{i=1}^N \mathcal{L}_{i_t}),  \\
\mathcal{P} &= 0.2 \frac{1}{N} \sum_{i=1}^N \frac{v_i}{u_i + v_i} +0.8 \frac{1}{N} \sum_{i=1}^N f(PMV) 
 \end{split}
\end{equation}

\item \textbf{No subgoals using $1$ DQN with $3$ inputs:} The setpoint is calculated using a single model of 1 DQN structured as Figure~\ref{fig:DQN}. The inputs are the three values of the fairness state ($s_t = (\mathcal{L}_1, \mathcal{L}_2, \mathcal{L}_3)$) \emph{without} the extra $4^{th}$ input. 
The reward is the same as Equation~\ref{eq:reward1dqn}.

\item \textbf{FaiRIoT~\cite{elmalaki2021fair}:} The closest to our approach is FaiRIoT which uses hierarchical RL.

\end{itemize}

To initialize the values of $s_t$ and the weights in the respective DQNs, we use the RR in the first $1200$ samples. 
To compare these methods, we investigated multiple evaluation metrics to evaluate the fairness across these three rooms; 1) the values of the fairness state ($s_t = (\mathcal{L}_1, \mathcal{L}_2, \mathcal{L}_3$), 2) a satisfaction metric, and the 3) PMV.
\vspace{-2mm}
\subsubsection{\textbf{Fairness state results}}


As explained in Section~\ref{sec:fairnessstate}, the best fairness state should be $s_t = (\mathcal{L}_1, \mathcal{L}_2, \mathcal{L}_3) = (1,1,1)$. To measure $s_t$, we need to use the satisfaction history counters by updating them per to Equation~\ref{eq:updatecounter}. In this application, we set the threshold $\tau$ to be $2.5$.  
Figure~\ref{fig:fsbigfigure} shows the fairness states results with $15$k samples equivalent to $62.5$ simulated days. In Figure~\ref{fig:fsbigfigure}-first row, we plot the values $\mathcal{L}_1$ for Room 1(red), $\mathcal{L}_2$ for Room 2(blue), and $\mathcal{L}_3$ for Room 3(green). \sysname achieves $s_t = ( \approx0.9991, \approx0.9989, \approx0.9991)$ after all the $3$ DQN running in the $3$ options converges at $t_s \approx 10k$. 
As for the Average approach $s_t$ stays approximately at $(\approx0.9957,\approx0.9962, \approx0.9982 )$, and for the RR approach $s_t$ stays at ($\approx0.9963,\approx 0.9932, \approx0.9924$).

To better get insights on how the different methods compare with respect to the values of $s_t$, in every time sample, we report the maximum value of $\mathcal{L}$ in $s_t$ (Figure~\ref{fig:fsbigfigure}-second row) and the minimum value of $\mathcal{L}$ in $s_t$ (Figure~\ref{fig:fsbigfigure}-third row). The value of $\mathcal{L}_i$ gives us an indication of how close Room \#$i$ is to the other rooms with respect to the satisfaction history records $\mathbf{c}$ as explained in Equation~\ref{eq:state}.


The results for using only 1 DQN with 3 and 4 inputs are unstable, and both have fluctuations in the fairness state values. This result is expected for 1 DQN since the fairness goal was not divided into sub-goals, unlike what the options framework provided. For space limitation, we only plot the results of 1 DQN with 3 inputs. However, the 1 DQN with 4 inputs had comparable results.

\vspace{-1mm}
\subsubsection{\textbf{Compare with group fairness definitions}} \label{sec:groupfairness}

\begin{figure*}[!t] 
\includegraphics[trim={0cm 8.5cm 0 0.1cm},clip,width=1\linewidth]{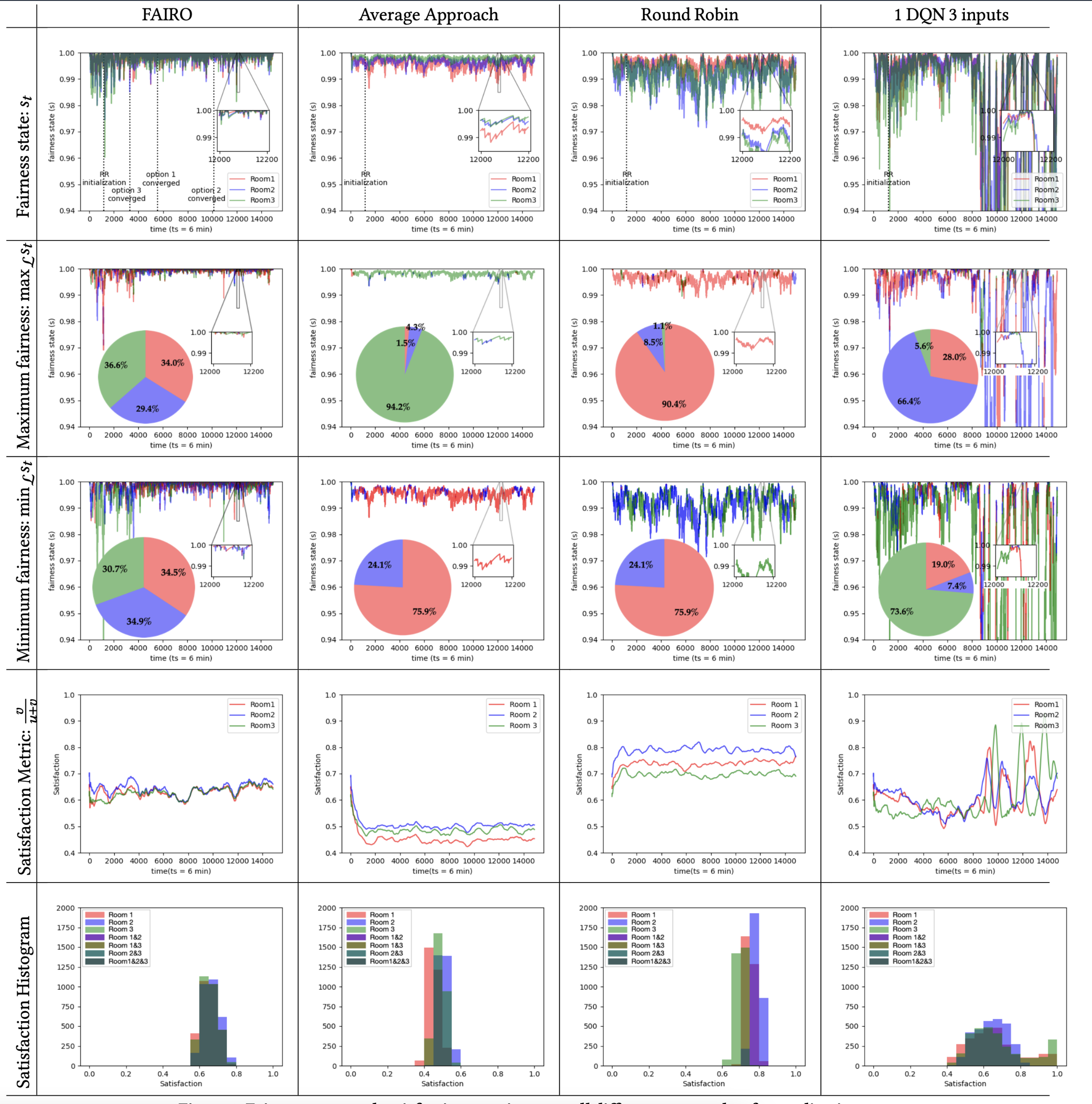}
\caption{Fairness state and satisfaction metric across all different approaches for application 1.}\vspace{-3mm}
\label{fig:fsbigfigure}
\end{figure*}

In Section~\ref{sec:rl}, we discussed the related work and highlighted the commonly used metrics for group fairness, namely, \emph{equal opportunity} and \emph{equalized odds}. Although these metrics are typically applied to binary classification tasks, we adopt their original definitions to evaluate and compare \sysname with the other approaches. In the context of group fairness, \emph{equal opportunity} aims to ensure that individuals from different groups have an equal chance or probability of experiencing positive outcomes or receiving beneficial treatment or resources. Therefore, to assess the performance of \sysname, we will analyze the results presented in Figure~\ref{fig:fsbigfigure} by examining the reported $\max_\mathcal{L} s_t$ values. Specifically, we will compare the probabilities of different rooms 
having the highest $\max_\mathcal{L} s_t$ values. For \emph{equal opportunity}, these probabilities need to be close, as denoted in Equation~\ref{eq:maxfair}. 
\setlength{\abovedisplayskip}{3pt}
\setlength{\belowdisplayskip}{3pt}
\begin{equation}\label{eq:maxfair}
 p(Room_i == \texttt{find}(\max_\mathcal{L}\ s_t)) \approx p(Room_{ N\text{\textbackslash}i} == \texttt{find}(\max_\mathcal{L}\ s_t) )\\
\end{equation}

As observed in Figure~\ref{fig:fsbigfigure}-second row, these probabilities are $34\%$, $29.4\%$, and $36.6\%$ for Room $\#1, \#2$, and $\# 3$ respectively, which are closer in values compared with the other approaches. 
In particular, using \sysname, the average absolute difference between the probabilities of \emph{equal opportunity} across the 3 rooms is reduced by $57.0\%$, $54.8\%$, $25.8\%$, and $35.8\%$ from Average Approach, RR, 1 DQN 4 inputs and 1 DQN 3 inputs respectively. \textbf{Accordingly, across all the approaches, \sysname improves the fairness by $43.35\%$ on average.}

While \emph{equal opportunity} focuses on the balance of the positive outcomes between groups, \emph{equalized odds} focuses on the balance between positive and negative outcomes across different groups. Hence, for the negative outcomes, we can examine the probabilities of different rooms having the $\min\mathcal{L} s_t$ values and assess whether they are approximately equal.
\begin{equation}\label{eq:minfair}
     p(Room_i == \texttt{find}(\min_\mathcal{L}\ s_t)) \approx p(Room _{N\text{\textbackslash}i} == \texttt{find}(\min_\mathcal{L}\ s_t) )
\end{equation}
As observed in Figure~\ref{fig:fsbigfigure}-third row, these probabilities are $34.5\%, 34.9\%$, and $30.7\%$ for Room $\#1, \#2$, and $\# 3$ respectively, which are closer in values compared with the other approaches. In particular, using \sysname the average absolute difference between the probabilities of \emph{equalized odds} across the 3 rooms is reduced by $47.8\%$, $35.8\%$, $35.5\%$, and $41.3\%$ from Average Approach, RR, 1 DQN 4 inputs, and 1 DQN 3 inputs respectively.


\begin{table*}[t!]
  \small
  \centering
  \begin{tabularx}{0.98\textwidth} {|p{0.12\textwidth}|p{0.12\textwidth}|p{0.12\textwidth}|p{0.12\textwidth}|p{0.12\textwidth}|p{0.12\textwidth}|p{0.12\textwidth}|}
    \hline
\textbf{Histogram} & \textbf{Rooms} & \textbf{\sysname} & \textbf{Average} & \textbf{RR} & \textbf{1 DQN 4 inputs} & \textbf{1 DQN 3 inputs} \\
\midrule
\multirow{3}{*}{\shortstack[l]{Satisfaction \\ Gaussian's fitting \\($\mu, \sigma^2) $} }& Room $1$   &  ( 0.642,0.047 )  & (0.452,0.030)  &   (0.748,0.021) & (0.629,0.066) & (0.657,0.142) \\
& Room $2$   &   (0.649, 0.050) & (0.506, 0.027)  &   (0.785,0.023) & (0.628, 0.076) & (0.641,0.084) \\
& Room $3$   &   (0.641,0.047) & (0.487,0.030)  &   (0.698,0.024) & (0.598,0.063) & (0.680,0.152) \\ \hline


 
\multirow{3}{*}{Satisfaction JSD} & Room $1\&2$  &   0.022 bits & 0.399 bits  &   0.269 bits & \textbf{0.015} bits & 0.115 bits\\
 & Room $1\&3$   &   \textbf{0.002} bits & 0.175 bits  &	0.443 bits & 0.049 bits & 0.020 bits\\
 & Room $2\&3$   &   \textbf{0.014} bits & 0.080 bits  &	0.835 bits & 0.041 bits & 0.121 bits \\ \cline{2-7}
 & JSD Average & \textbf{0.013} bits & 0.218 bits & 0.516 bits & 0.035 bits & 0.085 bits \\ \hline

    \bottomrule

\hline
\multirow{3}{*}{\shortstack[l]{PMV \\ Gaussian's fitting \\($\mu, \sigma^2) $} } & Room $1$   &   (0.346,0.490) & (0.369,0.430 )  &   (0.367, 0.590) & (0.353,0.459) & (0.336,0.502) \\
& Room $2$   &   (-0.014,0.612) & (0.021,0.498)  &   (0.021, 0.701) & (-0.003,0.563) & (-0.030,0.642) \\
& Room $3$   &    (-0.099,0.673) & (-0.065,0.564)  &   (-0.052,0.732) & (-0.095,0.647) & (-0.120,0.723) \\ \hline

\multirow{3}{*}{PMV JSD} & Room $1\&2$  &   0.088 bits & 0.112 bits  &   \textbf{0.071} bits & 0.092 bits & 0.089 bits\\
 & Room $1\&3$   &   0.119 bits & 0.152 bits  &	\textbf{0.099} bits & 0.124 bits & 0.119 bits\\
 & Room $2\&3$   &   0.012 bits & 0.023 bits  &	\textbf{0.014} bits & 0.014 bits & 0.014 bits \\ \cline{2-7}
 & JSD Average & 0.073 bits & 0.096 bits & \textbf{0.063} bits & 0.077 bits & 0.074 bits \\ \hline
    
  \end{tabularx}
  \caption{Comparison of the satisfaction values $\frac{v}{u+v}$ and the PMV across all the approaches in application 1. }\vspace{-4mm}
  \label{tbl:histsatpmv}
\end{table*}

\vspace{-2mm}
\subsubsection{\textbf{Satisfaction performance}}

In Equation~\ref{eq:rewardoptionapp1}, we used $\frac{v}{v+u}$ as one of the performance measures. Figure~\ref{fig:fsbigfigure}-fourth row shows the satisfaction values for all methods. \sysname achieves close satisfaction values across the three rooms over time. Average and RR approaches have stable but different satisfaction values across rooms while 1 DQN-3 inputs and 1-DQN-4 inputs show more fluctuations per room. We focused on samples after \sysname convergence ($12k$ to $15k$) and examined the satisfaction value histograms. Table~\ref{tbl:histsatpmv} reports the Jensen-Shannon Divergence (JSD) of these histograms\footnote{The Jensen–Shannon divergence is a method of measuring the similarity between two probability distributions~\cite{lin1991divergence}. The JSD is symmetric and always non-negative, with a value of $0$ indicating that the two distributions are identical, and a value greater than $0$ indicating that the two distributions are different.}. \sysname has the lowest average JSD ($0.013$), indicating closer satisfaction values across rooms. The mean value ($\mu$) of the satisfaction histogram fitting into a Gaussian distribution is approximately $0.64$ for all rooms. 
In particular, compared with other methods, \sysname JSD is reduced by $94.0\%$ from Average Approach, reduced by $97.5\%$ from RR, reduced by $62.9\%$ from 1 DQN 4 inputs method, and reduced by $84.7\%$ from 1 DQN 3 inputs method. Satisfaction histogram results indicate high similarity in satisfaction across rooms, reflecting fairness in adaptation actions over time. However, \sysname has a lower mean satisfaction per room compared to the RR, as fairness was an objective in the adaptation. RR has higher values since every time step one of the rooms can have its desired temperature which contributes to the number of samples with $100\%$ satisfaction. 

\vspace{-2mm}
\subsubsection{\textbf{PMV results}}
We report the statistics of the PMV across all approaches in Table~\ref{tbl:histsatpmv}.
\sysname achieves the second lowest average JSD and a comparable variance on PMV results, which indicates \sysname can improve fairness without hurting the PMV performance. The RR can achieve the lowest PMV JSD average since every time step one of the rooms can get exactly its desired temperature. Hence, the $3$ rooms can get almost identical PMV performance in the long term. In summary, \sysname's PMV JSD is reduced by $24.0\%$ from Average Approach, increased by $15.9\%$ from RR, reduced by $5.2\%$ from 1 DQN 4 inputs method, and reduced by $1.4\%$ from 1 DQN 3 inputs method.

\begin{table}[!t]
    \centering
    \vspace{-2mm}
\begin{tabularx}{\linewidth}{XX}
\includegraphics[width=\linewidth]{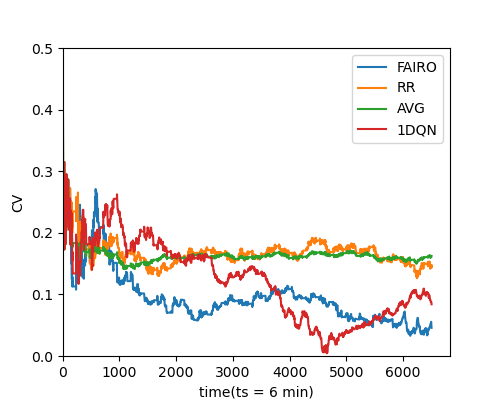}
    \captionof{figure}{Coeff. of variation ($cv$) of the temperature differences.}\label{fig:cvtdiff}  
& \includegraphics[width=\linewidth]{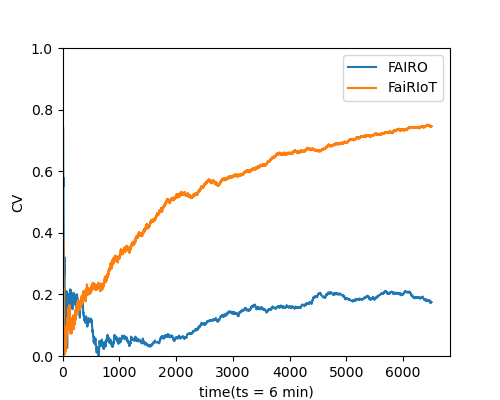}
     \captionof{figure}{Coeff. of variation ($cv$) of the utility $u_t$ over $\mathbf{w}$.}\label{fig:cvw} 
\end{tabularx}
\vspace{-3mm}
\label{fig:cv_w_tdiff}
\end{table}

\subsubsection{\textbf{Comparison with the State-of-the-Art FaiRIoT~\cite{elmalaki2021fair}}}
FaiRIoT uses a notion of utility $u_{h_t} = \frac{1}{t} \sum_{j=0}^{t} \frac{j}{t} w_{h_j} $ which is the average weight assigned by a layer called ``Mediator RL'' for a particular human $h$ over a time horizon $[0:t]$, where the factor $\frac{j}{t}$ is used to give more value to the recent weights learnt by the Mediator RL over the ones in the past. FaiRIoT measures the fairness of the Mediator RL using the coefficient of variation ($cv$) of the human utilities: $cv = \sqrt{\frac{1}{n-1} \sum_{h=1}^{n} \frac{(u_h - \bar{u})^2}{\bar{u}^2}},$ where $\bar{u}$ is the average utility of all humans. The Mediator RL is said to be more fair if and only if the $cv$ is smaller. Accordingly, we compare the $cv$ in FaiRIoT and \sysname in Figure~\ref{fig:cvw}. 
\sysname achieves $cv$ around $0.15$, while FaiRIoT $cv$ is larger than $0.6$. \textbf{Hence, \sysname improves the fairness where  $\mathbf{\emph{cv}}$ is reduced by $\mathbf{75\%}$.}

%% file: 07_application2Water.tex
\vspace{-2mm}
\section{Type 2: Water supply application} \label{sec:evalwater}

In the context of global climate change and rapid population growth, the planning and management of regional water systems play a crucial role~\cite{futureofwater, watercrisis}. 
As a result, Water Demand Models (WDMs) have been developed over the past few decades. These models serve two main purposes: gaining insights into water consumption behavior and forecasting demand. WDMs provide valuable inputs for decision-making processes related to the water distribution system (WDS) operation policies and infrastructure planning. 

Water authorities and governments have developed multiple management solutions to the water shortage problem ~\cite{jorgensen2009household}. However, it is a challenge to allocate water resources if the resource is always limited while providing every household with a fair experience on water resource distribution. Several works in the literature consider pricing and non-pricing policies in water management~\cite{jackson2005motivating, iglesias2008new, sechi2013water}. 
To evaluate \sysname for application Type 2, we assume that WDM is available to estimate the desired water demand per three households and the shared water resource to supply these different households is insufficient and time-varying~\cite{watercrisis}.

\vspace{-2mm}
\subsection{Household-Water Demand Model}
\begin{figure}[!t]
\centering
\includegraphics[width=0.6\columnwidth]{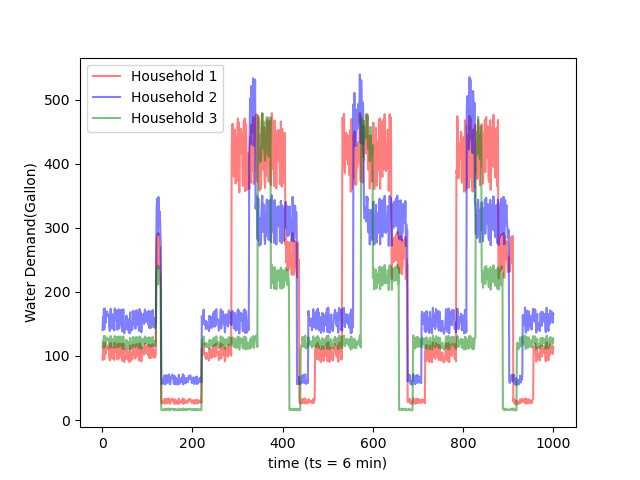}
\caption{Water demand patterns.}
\label{fig:watall}
\end{figure}

We used a WDM that supplies three households based on their residents' activity patterns~\cite{watermojtaba,water1mojtaba}. We used the same activity pattern in Section~\ref{sec:househumanmodel} and mapped them to water demand behavior as seen in Figure~\ref{fig:watall} which illustrates the water demand patterns per gallon for three households during three days. Every household has a tank for water reservation $\mathcal{T}$. The desired action for every household $h_i$ is to satisfy the water demand $d_i$.
Water resource ($\mathcal{R}_t$) is not fixed and insufficient to satisfy all demands. Since we assume that the WDM is available, we can assume that $\mathcal{R}_t$ can follow the demands profiles by choosing it to be $1.5$ the maximum demand of all the three demand profiles. Households' water demand profiles are independent of each other due to different human activities. We extended the Mathworks Water Supply model to include the household water demand profiles and multiple houses with a limited shared water supply resource~\cite{MATLABwater}.

\vspace{-2mm}
\subsection{Context-aware engine}
To focus on the main contribution and not designing a new water consumption behavior or forecasting demand models, the context-aware engine, as shown in Figure~\ref{fig:framework} provides the desired action for every household $h_i$, which is the current water demand $d_i$ based on the water demand pattern.
This demand is supposed to be satisfied via the water supply $s_i$ from the shared limited resource $\mathcal{R}_t$ and the current reserved water in the household water tank $t_i$. Hence, we define the application performance as the percentage of balancing the demand and the supply.
\setlength{\abovedisplayskip}{3pt}
\setlength{\belowdisplayskip}{3pt}
\begin{equation}\label{eq:balancerate}
     \text{Balance Rate (BR)} = \frac{ \text{supply } s_i + \text{reserve } t_i}{ \text{demand } d_i}
\end{equation}





\vspace{-4mm}
\subsection{Evaluation}\label{sec:Evaluationapp2}
We compare the same methods as in the first application. 

\begin{itemize}[leftmargin=*, noitemsep, topsep=0pt]
\item \textbf{\sysname:} The supply $s_i$ for each household $h_i$ is calculated by the \sysname algorithm. In particular, $s_i$ receives $w_i \mathcal{R}_t$ as explained in Section~\ref{sec:apptypes} where the global action $a_{g_t}$ is the weighted distribution of this resource $\mathcal{R}_t$. The reward is the same as explained in the first application in Equation~\ref{eq:rewardoptionapp1} using performance $\mathcal{P}_i$ based on the BR. Hence, $\mathcal{P}_i = 0.2 \frac{v_i}{u_i+v_i} + 0.8 f(\text{BR})$.

\item \textbf{Average approach:} Each household $h_i$ has the same amount of supply. Hence, $s_{i} = \frac{1}{3}\ \mathcal{R}_t $. We further augment the average approach in this application by using a \textbf{Weighted Average approach}. In particular, each household supply is proportional to its demand. Hence, $s_{i} = \frac{d_i}{\sum_{i=1}^3 d_i} \mathcal{R}_t $.

\item \textbf{Round Robin (RR):}
Each time step, one of the households in a rotation will be guaranteed sufficient supply to cover its demand. Leftovers from the resource will be shared equally with all households. For example, at time $t=1$, $s_1 = d_1$ while $s_2 = s_3 = \frac{\mathcal{R}_t - d_1}{2}$. 
We further augmented the RR to consider a \textbf{Weighted RR}.
In this case, supplies are calculated similarly to RR, but the leftover water will be distributed proportionally to their demands as in the weighted average approach.

\item \textbf{No subgoals using $1$ DQN with $4$ inputs:} Supply for each household is calculated by 1 DQN structure as explained in Section~\ref{sec:Evaluationapp1}.

\item \textbf{No subgoals using $1$ DQN with $3$ inputs:} Supply for each household is calculated by 1 DQN structure as explained in Section~\ref{sec:Evaluationapp1}.
\end{itemize}

To compare these methods, we investigated multiple evaluation metrics to evaluate the fairness across these three rooms; 1) the fairness state ($s_t$) 
2) a satisfaction metric, and 3) the balance rate (BR).

\begin{figure*}[!t]
\includegraphics[trim={0cm 8.4cm 0 0},clip,width=1\linewidth]{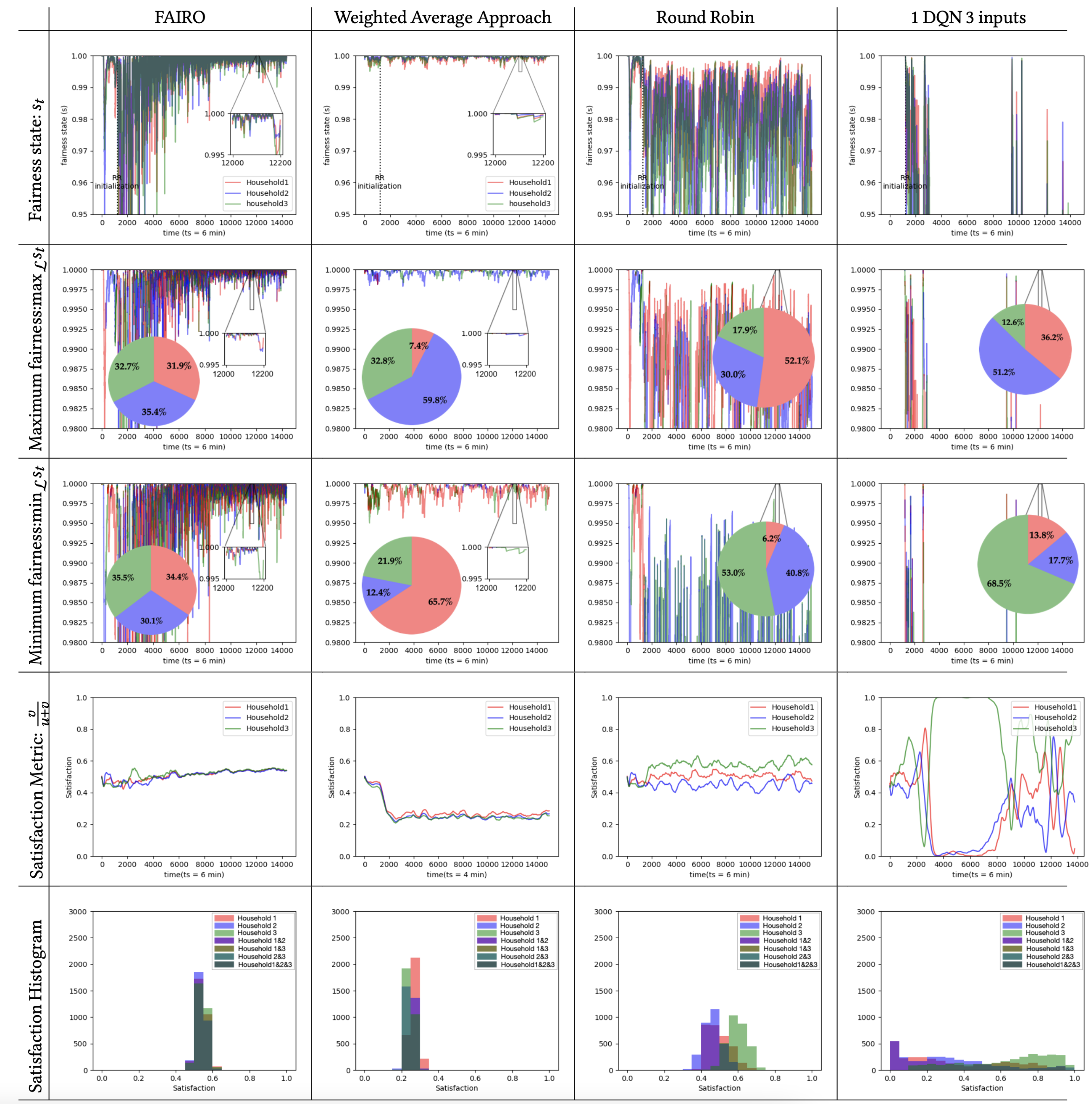}
\caption{Fairness state and satisfaction metric across all different approaches for application 2.
}\vspace{-3mm}
\label{fig:fsbigfiguretank}
\end{figure*}

\begin{table*}[!h]
  \small
  \centering
  \begin{tabularx}{1.\textwidth}{|p{1.9cm}|p{1.9cm}||p{1.5cm}|p{1.5cm}|p{1.5cm}|p{1.5cm}|p{1.5cm}|p{1.5cm}|p{1.5cm}|}
    \hline
\textbf{Histogram} & \textbf{Rooms} & \textbf{\sysname} & \textbf{Weighted Average}& \textbf{Weighted RR} & \textbf{Average} & \textbf{RR} & \textbf{1 DQN 4 inputs} & \textbf{1 DQN 3 inputs}\\
\midrule
\multirow{3}{*}{\shortstack[l]{Satisfaction \\ Gaussian's fitting \\($\mu, \sigma^2) $} }  & Room $1$   &   (\textbf{0.540}, 0.037) & (0.267, 0.021)  & (0.459, 0.014) & (0.214, 0.100)  & (0.496, 0.061) & (0.475, 0.254)  & (0.342, 0.280) \\
 & Room $2$   &   (\textbf{0.540}, 0.029) & (0.249, 0.022)  &   (0.445, 0.014) & (0.205, 0.085)  & (0.462, 0.050) & (0.339, 0.190)  & (0.315, 0.257) \\
& Room $3$   &    (0.535, 0.048) & (0.244, 0.020)  & (0.429, 0.013) & (0.495, 0.105)  & (0.596, 0.052) & (0.493, 0.155)  & (\textbf{0.636}, 0.241) \\ \hline
\multirow{3}{*}{\shortstack[l]{Satisfaction \\ JSD}} & Room $1\&2$  &   \textbf{0.016} bits & 0.091 bits  &   0.079 bits & 0.088 bits   & 0.094 bits & 0.123 bits & 0.048 bits\\
 & Room $1\&3$   &   \textbf{0.011} bits & 0.150 bits  &	0.427 bits & 0.654 bits   & 0.390 bits & 0.047 bits & 0.237 bits\\
 & Room $2\&3$   &   0.024 bits & \textbf{0.010} bits  &	0.165 bits  & 0.756 bits & 0.644 bits & 0.135 bits & 0.288 bits \\ \cline{2-9}
 & JS Average & \textbf{0.017} bits & 0.084 bits & 0.224 bits  & 0.499 bits  & 0.376 bits & 0.102 bits & 0.191 bits \\ \hline \hline
\multirow{1}{*}{\shortstack[l]{Balance Rate (BR)}} & avg. >80\% BR
    & \textbf{0.535(1606)} & 0.254(763) & 0.446(1336.7) & 0.305(915.3) & 0.517(1552) & 0.432(1295.7) & 0.432(1296.3) \\\cline{2-9}
    \hline

  \end{tabularx}
  \caption{Comparison of the satisfaction values $\frac{v}{u+v}$ and the balance rate across all the approaches in application 2. \vspace{-4mm}
  \label{tbl:app2table}}
\end{table*}

\vspace{-2mm}
\subsubsection{\textbf{Fairness state results}}
We report in Figure ~\ref{fig:fsbigfiguretank} the fairness states results with $15k$ samples equivalent to $62.5$ simulated days. 
To measure $s_t$, we need to use the satisfaction history counters by updating them per Equation~\ref{eq:updatecounter}. In this application, we set the $\| d_i - (s_i + t_i)\| \leq \tau$ with $\tau$ equals $20\%$ of the demand $d_i$ which means BR is $80\%$. 

Figure ~\ref{fig:fsbigfiguretank}-first row plot the fairness state results for each household. \sysname converges at $t_s\approx10k$. 
The results of 1 DQN 3 inputs and 1 DQN 4 inputs are unstable with a lot of fluctuations. 

\vspace{-2mm}
\subsubsection{\textbf{Compare with group fairness definitions}} 
Figure~\ref{fig:fsbigfiguretank}-second and third rows show the maximum and minimum value of $s_t$. 
\sysname achieves \emph{equal opportunity} probabilities $31.9\%$, $35.4\%$, and $32.7\%$ for household $\#1$, $\#2$, and $\#3$ respectively, which are closer in values than other methods. In particular, using \sysname, the average absolute difference between the probabilities of \emph{equal opportunity} across the 3 households is reduced by $32.6\%$, $43.5\%$, $27.7\%$, $20.5\%$, $10.6\%$, and $23.4\%$ from Weighted Average, Weighted RR, Average, RR, 1DQN 4 inputs and 1 DQN 3 inputs respectively. \textbf{Accordingly, across all the approaches, \sysname improves the fairness by $26.38\%$ on average.} As for \emph{equalized odds} probabilities, \sysname reports $34.4\%$, $30.1\%$, and $35.5\%$ for households $\#1, \#2$, and $\# 3$ respectively, which are also closer in values compared to the other approaches. In particular, using \sysname, the average absolute difference between the probabilities of \emph{equalized odds}  across the 3 households is reduced by $31.9\%$, $31.0\%$, $37.1\%$, $27.6\%$, $9.4\%$, and $32.9\%$ from Weighted Average, Weighted RR, Average, RR, 1DQN 4 inputs and 1 DQN 3 inputs respectively.

\vspace{-2mm}
\subsubsection{\textbf{Satisfaction performance}}
As explained in Equation ~\ref{eq:rewardoptionapp1}, the value $\frac{v}{v+u}$ is used as a term to measure the performance. Figure ~\ref{fig:fsbigfiguretank}-forth row reports the satisfaction values across the three households. Satisfaction values in \sysname are closer compared with other approaches with satisfaction values around $0.55$. RR satisfaction values are from $0.6$ to $0.4$ with different between households. The Weighted Average Approach satisfaction values are close, but values stay below $0.3$. 
1 DQN-3 inputs and 1 DQN-4 inputs have unstable fluctuations across $3$ households. 

We use $3k$ samples after \sysname convergence($12k$ to $15k$) to examine the satisfaction value histograms. Table~\ref{tbl:app2table} reports the Jensen-Shannon Divergence (JSD) of these histograms. \sysname has the lowest average JSD ($0.017$), indicating closer satisfaction values across rooms. 
In particular, compared with other methods, \sysname JSD is reduced by $79.8\%$, $92.4\%$, $96.5\%$, $95.5\%$, $83.3\%$, and $91.1\%$ from Weighted Average Approach, Weighted RR, Average Approach, RR, 1 DQN 4 inputs, and 1 DQN 3 inputs methods respectively. 


\vspace{-2mm}
\subsubsection{\textbf{Balance rate (BR) results}}
We report the average number of samples that have $>80\%$ BR across all households in brackets in Table~\ref{tbl:app2table} and average it over 3k samples. Samples have $>80\%$ BR correspond to satisfactory samples. Table~\ref{tbl:app2table}-last row shows \sysname has the highest average $>80\%$ BR value(0.535). In particular, compared with other methods, \sysname's average $>80\%$ BR is improved by $110.6\%$, $20.0\%$, $75.4\%$, $3.5\%$, $23.8\%$, and $23.8\%$ from Weighted Average Approach, Weighted RR, Average Approach, RR, 1 DQN 4 inputs, and 1 DQN 3 inputs methods respectively.

%% file: 08_application3Learning.tex
\section {Type 3: Smart learning system}\label{sec:evallearning}
Monitoring the human learning state and performance is crucial for assessing progress, identifying gaps, and personalizing instruction
~\cite{terai2020detecting}. 
With the wide adoption of immersive technology, such as virtual reality (VR) in education environments, recent studies showed that these technologies would have a significant impact on the learning~\cite{ibanez2014experimenting}, and workforce training~\cite{irizarry2013infospot}. However, during elongated education periods, especially in an online or VR setup, human performance is prone to significantly decline~\cite{terai2020detecting} due to distractions, drowsiness, and fatigue. In this application, we examine a setup where multiple humans share the same educational environment, and a HITL learning system adapts this environment by enabling an immersive VR experience to improve the learning experience.  

\vspace{-2mm}
\subsection{Human-Learning VR model}\label{sec:humanVRmodel}
\begin{figure}[!t]
\centering
\includegraphics[width=1\columnwidth]{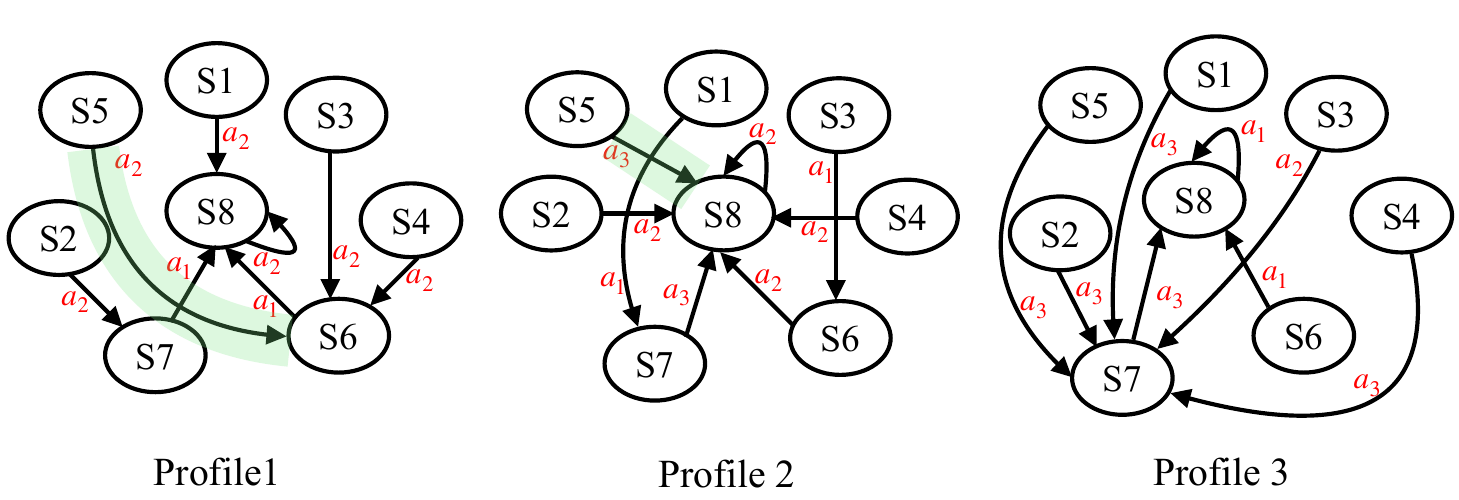}
\caption{MDP for three human profiles in VR learning environment.}
\label{fig:profiles}\vspace{+1mm}
\end{figure}

We used the dataset associated with recent work in the literature that examined the effect of using VR on the learning performance of $15$ participants watching lectures from Khan Academy~\cite{taherisadr2023adaparl,khan}. In particular, the human learning experience can be determined through three main features, including alertness level, fatigue level, and vertigo level~\cite{brunnstrom2020latency}, forming a total of $8$ states if we use binary values (e.g., Alert:1, Not alert:0) for these features. Indeed different humans may transition between these states based on their experience and preference for VR technology. A HITL learning environment monitors these states and provides the best adaptation action to provide a better learning experience regarding the human state. These adaptation actions can (1) $a_1$ give the human a small break, (2) $a_2$ enable the use of VR, or (3) $a_3$ disable VR. 
Accordingly, we used this dataset to categorize the $15$ human behavior into three groups. The first group profile has humans who are most tolerant to VR, meaning they do not experience simulator sickness after using VR devices for more than $20$ minutes, the second group profile has humans who can show some cybersickness during VR exposure, and the third group profile has the humans with the least VR tolerance. We model these groups' behavior using MDP as shown in Figure~\ref{fig:profiles}. In particular, state $8$ encodes the best human state regarding high alertness, vigor, and no sense of vertigo or dizziness. In contrast, state $1$ encodes the worst human state. 
Based on these three human models, we instantiated three humans, one from each profile to run a synthesized experiment of three humans having a class schedule every day. At the beginning of every day, the states of the three humans are initialized to state 3 or state 1 randomly. 

\begin{figure*}[!th]
\includegraphics[trim={0cm 4.8cm 0 0},clip,width=1\linewidth]{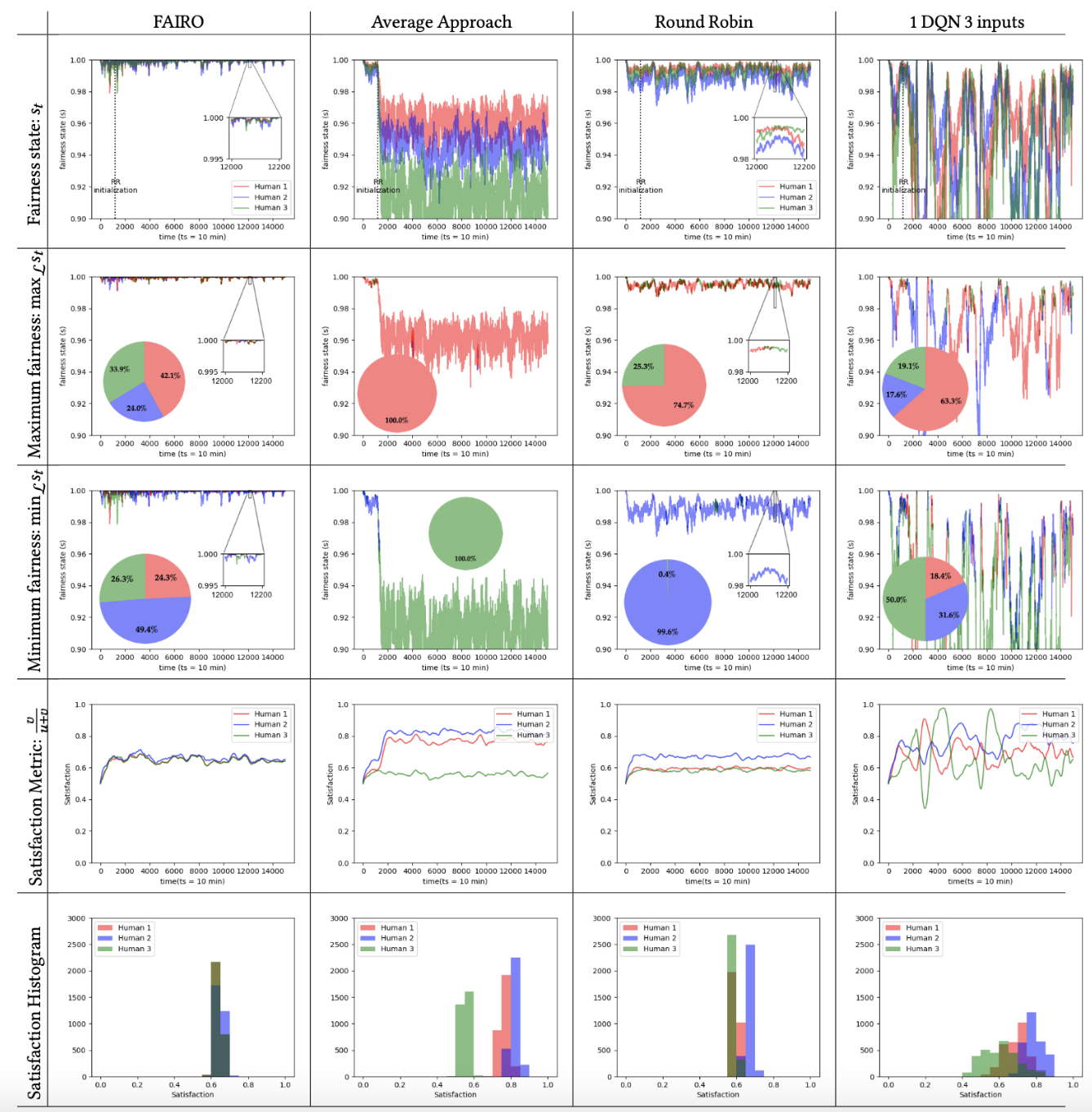}
\caption{Fairness state and satisfaction metric across all different approaches for application 3.}\vspace{-3mm}
\label{fig:fsbigfigureapp3}
\end{figure*}

\begin{table*}[!h]
  \small
  \centering
  \begin{tabularx}{0.99\textwidth} {|p{0.17\textwidth}|p{0.12\textwidth}|p{0.11\textwidth}|p{0.11\textwidth}|p{0.11\textwidth}|p{0.11\textwidth}|p{0.11\textwidth}|}
    \hline
\textbf{Histogram} & \textbf{Rooms} & \textbf{\sysname} & \textbf{Average} & \textbf{RR} & \textbf{1 DQN 4 inputs} & \textbf{1 DQN 3 inputs} \\
\midrule

\multirow{3}{*}{\shortstack[l]{Satisfaction \\ Gaussian's fitting \\$(\mu, \sigma^2) $} }& Room $1$   &  (0.639, 0.017)  & (0.761, 0.033)  &   (0.594, 0.015) & (0.708, 0.053) & (0.692, 0.062) \\
& Room $2$   &   (0.647, 0.020) &  (0.821, 0.020)  &  (0.671, 0.018) & (0.811, 0.057) & (0.787, 0.050) \\
& Room $3$   &   (0.638, 0.018) & (0.554, 0.020)  &   (0.583, 0.012) & (0.638, 0.115) & (0.604, 0.094) \\ \hline

\multirow{3}{*}{Satisfaction JSD} & Room $1\&2$  &   \textbf{0.022} bits & 0.528 bits  &   0.801 bits & 0.459 bits & 0.396 bits\\
 & Room $1\&3$   &   \textbf{0.000} bits & 1 bits  &	0.060 bits & 0.494 bits & 0.229 bits\\
 & Room $2\&3$   &   \textbf{0.021} bits & 1 bits  &	0.884 bits & 0.538 bits & 0.657 bits \\ \cline{2-7}
 & JS Average & \textbf{0.094} bits & 0.843 bits & 0.582 bits & 0.497 bits & 0.427 bits \\ \hline
Learning Experience (LE) & Overlap Samples \% & 0.488 & 0.438 & 0.507 & 0.473 & \textbf{0.497} \\ 
\hline

  \end{tabularx}
  \caption{Comparison of the satisfaction values $\frac{v}{u+v}$ and the learning experience (LE) in application 3. }\vspace{-4mm}
  \label{tbl:app3table}
\end{table*}

\vspace{-3mm}
\subsection{Context-aware engine} 
The desired action $d_i$ is selected to improve the human learning experience.
Unlike the previous two applications, this application has a non-numerical action space, as mentioned in Section~\ref{sec:humanVRmodel}. Hence, to use \sysname we need to quantify this categorical action in terms of its effect as explained in Section~\ref{sec:apptypes}. We exploit the MDP model in Figure~\ref{fig:profiles} to quantify these actions. In particular, as $S_8$ and $S_0$ encode the best and the worst state, respectively, we assign a value to each state linearly with $S_8$ as the highest value. Hence, the desired action $d_i$ that can make a transition to a better state in the MDP has a high numerical value. However, every human may be in a different state. Hence, for every desired action $d_i$, we calculate the \emph{effect} denoted as $k_i$ of applying $d_i$ on the \emph{other} humans. In particular, using the MDP models, if the desired action $d_i$ from human $h_i$ were to be applied in the shared environment, we check the state transition it will cause on all the other humans $h_{N \text{\textbackslash} i}$. The difference in the values of the two states that $d_i$ can cause the transition is used to measure its effect. In our setup, we have three humans; hence to measure the effect $k_1$ of applying desired action $d_1$ from the first human is the added values of the transition it can cause on human 2 and human 3 based on their current states and their MDP model.
Hence, after applying an adaptation action, the human learning experience can be measured as the improvement in the human state values and the current human state value.   
\begin{equation}\label{eq:humanexperience}
\text{Learning Experience (LE)} = \text{state improvement} + \text {state value}\ \in [-1 , 1]
\end{equation}

\vspace{-4mm}
\subsection{Evaluation}
We compare $5$ methods as in the first application. 


\begin{itemize}[leftmargin=*, noitemsep, topsep=0pt]
\item \textbf{\sysname:} 
The global action $a_{g_t} = d_i$ is the desired action of the human$_i$ with the maximum weighted effect as explained in Section~\ref{sec:apptypes}. The reward is the same as explained in the first application in Equation~\ref{eq:rewardoptionapp1} using performance $\mathcal{P}_i$ based on the learning experience (LE). Hence, $\mathcal{P}_i = 0.2 \frac{v_i}{u_i+v_i} + 0.8 f(\text{LE})$. 

\item \textbf{Average approach:} The global action $a_{g_t} = d_i$ is the desired action of the human$_i$ with the median weighted effect.

\item \textbf{Round Robin (RR):} The global action $a_{g_t}$ is selected from one of the desired actions of all humans in a rotation.

\item \textbf{No subgoals using 1 DQN with 3 inputs and 1 DQN 4 inputs:} The weighted effects are determined by using 1 DQN structure (as explained in Section ~\ref{sec:Evaluationapp1}) and then the global action is the desired action with the maximum weighted effect.
\end{itemize}

To compare these methods, we investigated multiple evaluation metrics; 1) the values of the fairness state, 2) a satisfaction metric, and 3) learning experience (LE).

\vspace{-2mm}
\subsubsection{\textbf{Fairness state results}}
Figure ~\ref{fig:fsbigfigureapp3} reports the fairness states results with $15k$ samples. To measure $s_t$, we need to use the satisfaction history counters by updating them per Equation~\ref{eq:updatecounter}. In this application, if the learning experience (LE) of the human is $>0$ as measured in Equation~\ref{eq:humanexperience}, it will be considered satisfied. 
Figure ~\ref{fig:fsbigfigureapp3}-first row plots the fairness state results for 3 humans. \sysname converges at $t_s\approx10k$. 

\vspace{-2mm}
\subsubsection{\textbf{Compare with group fairness definitions}}
Figure~\ref{fig:fsbigfiguretank}-second and third rows show the maximum and minimum value of $s_t$. 
\sysname achieves \emph{equal opportunity} probabilities $42.1\%$, $24.0\%$, and $33.9\%$ for human $\#1$, $\#2$, and $\#3$ respectively. As for \emph{equalized odds} probabilities, \sysname reports $24.3\%$, $49.4\%$, and $26.3\%$ for humans $\#1, \#2$, and $\# 3$ respectively.
While the difference in these values across the three humans is not as small as we had in the other two applications, we observe that this difference in \sysname is better than the other approaches. In particular, the average absolute difference between the probabilities of \emph{equal opportunity} across the 3 humans is reduced by $54.6\%$, $37.7\%$, $18.5\%$, and $34.6\%$ from Average Approach, RR, 1 DQN 3 inputs, and 1 DQN 4 inputs respectively. \textbf{Accordingly, across all the approaches, \sysname improves the fairness by $36.35\%$ on average.} Similarly, the average absolute difference between the probabilities of \emph{equalized odds} across the 3 humans is reduced by $49.9\%$, $49.7\%$, $4.3\%$, and $18.3\%$ from Average Approach, RR, 1 DQN 3 inputs, and 1 DQN 4 inputs respectively.


\vspace{-2mm}
\subsubsection{\textbf{Satisfaction performance}}

Figure ~\ref{fig:fsbigfiguretank}-fourth row shows the satisfaction values ($\frac{v}{v+u}$) across three humans. \sysname results show the three human satisfaction values stay at $0.6$ and they are closer to each other compared with other approaches. We use $3k$ samples after \sysname convergence(12k to 15k) to examine the satisfaction value histograms. Table~\ref{tbl:app3table} reports the Jensen-Shannon Divergence (JSD) of these histograms. \sysname has the lowest average JS ($0.021$), indicating closer satisfaction values across rooms. The mean value ($\mu$) of the satisfaction histogram fitting into a Gaussian distribution is approximately $0.64$ for all rooms. 



\vspace{-2mm}
\subsubsection{\textbf{Learning experience (LE) results}}
We count the number of samples with positive LE and average over $3k$ samples.
As shown in Table~\ref{tbl:app3table}, \sysname achieves comparable LE results. \sysname LE is improved by $11.4\%$ from Average Approach, reduced by $3.4\%$ from RR, reduced by $1.8\%$ from 1 DQN 4 inputs, and reduced by $3.2\%$ from 1 DQN 3 inputs.

%% file: 09_conclusion.tex
\vspace{-2mm}
\section{Conclusion}
This paper solves the fairness task by dividing it into smaller subgoals over a time trajectory for fairness-aware sequential-decision making within HITL systems. We propose \sysname framework that considers the nuances of human variability and preferences over time. \sysname offers a powerful algorithm to address various application setups, facilitating equitable decision-making processes in HITL environments. 

%% file: 00_main.bbl

\begin{thebibliography}{70}


\ifx \showCODEN    \undefined \def \showCODEN     #1{\unskip}     \fi
\ifx \showDOI      \undefined \def \showDOI       #1{#1}\fi
\ifx \showISBNx    \undefined \def \showISBNx     #1{\unskip}     \fi
\ifx \showISBNxiii \undefined \def \showISBNxiii  #1{\unskip}     \fi
\ifx \showISSN     \undefined \def \showISSN      #1{\unskip}     \fi
\ifx \showLCCN     \undefined \def \showLCCN      #1{\unskip}     \fi
\ifx \shownote     \undefined \def \shownote      #1{#1}          \fi
\ifx \showarticletitle \undefined \def \showarticletitle #1{#1}   \fi
\ifx \showURL      \undefined \def \showURL       {\relax}        \fi
\providecommand\bibfield[2]{#2}
\providecommand\bibinfo[2]{#2}
\providecommand\natexlab[1]{#1}
\providecommand\showeprint[2][]{arXiv:#2}

\bibitem[wat(2023)]%
        {watercrisis}
 \bibinfo{year}{2023}\natexlab{}.
\newblock \bibinfo{booktitle}{\emph{The water crisis is worsening. Researchers
  must tackle it together}}.
\newblock
\urldef\tempurl%
\url{https://www.nature.com/articles/d41586-023-00182-2}
\showURL{%
\tempurl}


\bibitem[Academy(2023)]%
        {khan}
\bibfield{author}{\bibinfo{person}{Khan Academy}.}
  \bibinfo{year}{2023}\natexlab{}.
\newblock \bibinfo{title}{Khan Academy}.
\newblock \bibinfo{howpublished}{\url{https://www.khanacademy.org/}}.
\newblock
\newblock
\shownote{Accessed: 2023-05-07}.


\bibitem[Ahadi-Sarkani and Elmalaki(2021)]%
        {ahadi2021adas}
\bibfield{author}{\bibinfo{person}{Armand Ahadi-Sarkani} {and}
  \bibinfo{person}{Salma Elmalaki}.} \bibinfo{year}{2021}\natexlab{}.
\newblock \showarticletitle{ADAS-RL: Adaptive vector scaling reinforcement
  learning for human-in-the-loop lane departure warning}. In
  \bibinfo{booktitle}{\emph{Proceedings of the First International Workshop on
  Cyber-Physical-Human System Design and Implementation}}.
  \bibinfo{pages}{13--18}.
\newblock


\bibitem[Angwin et~al\mbox{.}({[n.\,d.]})]%
        {angwin2016machine}
\bibfield{author}{\bibinfo{person}{Julia Angwin}, \bibinfo{person}{Jeff
  Larson}, \bibinfo{person}{Surya Mattu}, {and} \bibinfo{person}{Lauren
  Kirchner}.} \bibinfo{year}{[n.\,d.]}\natexlab{}.
\newblock \bibinfo{title}{Machine Bias: There’s software used across the
  country to predict future criminals. And it’s biased against blacks.}
\newblock
  \bibinfo{howpublished}{\url{https://www.propublica.org/article/machine-bias-risk-assessments-in-criminal-sentencing}}.
\newblock


\bibitem[Annaswamy et~al\mbox{.}(2023)]%
        {george2023roadmap}
\bibfield{author}{\bibinfo{person}{Anuradha Annaswamy}, \bibinfo{person}{Karl
  Johansson}, {and} \bibinfo{person}{George Pappas}.}
  \bibinfo{year}{2023}\natexlab{}.
\newblock \bibinfo{title}{Control for Societal-Scale Challenges Roadmap 2030}.
\newblock
\newblock


\bibitem[{ASHRAE/ANSI Standard 55-2010 American Society of Heating,
  Refrigerating, and Air-Conditioning Engineers}(2010)]%
        {handbook2009american}
\bibfield{author}{\bibinfo{person}{{ASHRAE/ANSI Standard 55-2010 American
  Society of Heating, Refrigerating, and Air-Conditioning Engineers}}.}
  \bibinfo{year}{2010}\natexlab{}.
\newblock \showarticletitle{Thermal environmental conditions for human
  occupancy}.
\newblock \bibinfo{journal}{\emph{Inc.Atlanta, GA, USA}}
  (\bibinfo{year}{2010}).
\newblock


\bibitem[Avni et~al\mbox{.}(2015)]%
        {watermojtaba}
\bibfield{author}{\bibinfo{person}{Noa Avni}, \bibinfo{person}{Barak Fishbain},
  {and} \bibinfo{person}{Uri Shamir}.} \bibinfo{year}{2015}\natexlab{}.
\newblock \showarticletitle{Water consumption patterns as a basis for water
  demand modeling}.
\newblock \bibinfo{journal}{\emph{Water Resources Research}}
  \bibinfo{volume}{51}, \bibinfo{number}{10} (\bibinfo{year}{2015}),
  \bibinfo{pages}{8165--8181}.
\newblock


\bibitem[Bacon et~al\mbox{.}(2017)]%
        {bacon2017option}
\bibfield{author}{\bibinfo{person}{Pierre-Luc Bacon}, \bibinfo{person}{Jean
  Harb}, {and} \bibinfo{person}{Doina Precup}.}
  \bibinfo{year}{2017}\natexlab{}.
\newblock \showarticletitle{The option-critic architecture}. In
  \bibinfo{booktitle}{\emph{Proceedings of the AAAI conference on artificial
  intelligence}}, Vol.~\bibinfo{volume}{31}.
\newblock


\bibitem[Bassen et~al\mbox{.}(2020)]%
        {bassen2020reinforcement}
\bibfield{author}{\bibinfo{person}{Jonathan Bassen}, \bibinfo{person}{Bharathan
  Balaji}, \bibinfo{person}{Michael Schaarschmidt}, \bibinfo{person}{Candace
  Thille}, \bibinfo{person}{Jay Painter}, \bibinfo{person}{Dawn Zimmaro},
  \bibinfo{person}{Alex Games}, \bibinfo{person}{Ethan Fast}, {and}
  \bibinfo{person}{John~C Mitchell}.} \bibinfo{year}{2020}\natexlab{}.
\newblock \showarticletitle{Reinforcement Learning for the Adaptive Scheduling
  of Educational Activities}. In \bibinfo{booktitle}{\emph{Proceedings of the
  2020 CHI Conference on Human Factors in Computing Systems}}.
  \bibinfo{pages}{1--12}.
\newblock


\bibitem[Brunnstr{\"o}m et~al\mbox{.}(2020)]%
        {brunnstrom2020latency}
\bibfield{author}{\bibinfo{person}{Kjell Brunnstr{\"o}m},
  \bibinfo{person}{Elijs Dima}, \bibinfo{person}{Tahir Qureshi},
  \bibinfo{person}{Mathias Johanson}, \bibinfo{person}{Mattias Andersson},
  {and} \bibinfo{person}{M{aa}rten Sj{\"o}str{\"o}m}.}
  \bibinfo{year}{2020}\natexlab{}.
\newblock \showarticletitle{Latency impact on Quality of Experience in a
  virtual reality simulator for remote control of machines}.
\newblock \bibinfo{journal}{\emph{Signal Processing: Image Communication}}
  \bibinfo{volume}{89} (\bibinfo{year}{2020}), \bibinfo{pages}{116005}.
\newblock


\bibitem[Carroll(2007)]%
        {carrollpulmonary2007}
\bibfield{author}{\bibinfo{person}{Robert~G. Carroll}.}
  \bibinfo{year}{2007}\natexlab{}.
\newblock \showarticletitle{Pulmonary System}.
\newblock In \bibinfo{booktitle}{\emph{{Elsevier's Integrated Physiology}}}.
  \bibinfo{publisher}{Elsevier}, Chapter~10, \bibinfo{pages}{99--115}.
\newblock


\bibitem[Chouldechova and Roth(2018)]%
        {chouldechova2018frontiers}
\bibfield{author}{\bibinfo{person}{Alexandra Chouldechova} {and}
  \bibinfo{person}{Aaron Roth}.} \bibinfo{year}{2018}\natexlab{}.
\newblock \showarticletitle{The frontiers of fairness in machine learning}.
\newblock \bibinfo{journal}{\emph{arXiv preprint arXiv:1810.08810}}
  (\bibinfo{year}{2018}).
\newblock


\bibitem[Claure et~al\mbox{.}(2020)]%
        {claure2020multi}
\bibfield{author}{\bibinfo{person}{Houston Claure}, \bibinfo{person}{Yifang
  Chen}, \bibinfo{person}{Jignesh Modi}, \bibinfo{person}{Malte Jung}, {and}
  \bibinfo{person}{Stefanos Nikolaidis}.} \bibinfo{year}{2020}\natexlab{}.
\newblock \showarticletitle{Multi-armed bandits with fairness constraints for
  distributing resources to human teammates}. In
  \bibinfo{booktitle}{\emph{Proceedings of the 2020 ACM/IEEE International
  Conference on Human-Robot Interaction}}. \bibinfo{pages}{299--308}.
\newblock


\bibitem[Cohen et~al\mbox{.}(2019)]%
        {cohen2019efficient}
\bibfield{author}{\bibinfo{person}{Lee Cohen}, \bibinfo{person}{Zachary~C
  Lipton}, {and} \bibinfo{person}{Yishay Mansour}.}
  \bibinfo{year}{2019}\natexlab{}.
\newblock \showarticletitle{Efficient candidate screening under multiple tests
  and implications for fairness}.
\newblock \bibinfo{journal}{\emph{arXiv preprint arXiv:1905.11361}}
  (\bibinfo{year}{2019}).
\newblock


\bibitem[Council(2021)]%
        {futureofwater}
\bibfield{author}{\bibinfo{person}{National~Intelligence Council}.}
  \bibinfo{year}{2021}\natexlab{}.
\newblock \bibinfo{booktitle}{\emph{The Future of Water: Water Insecurity
  Threatening Global Economic Growth, Political Stability}}.
\newblock
\urldef\tempurl%
\url{https://www.dni.gov/index.php/gt2040-home/gt2040-deeper-looks/future-of-water}
\showURL{%
\tempurl}


\bibitem[Creager et~al\mbox{.}(2020)]%
        {creager2020causal}
\bibfield{author}{\bibinfo{person}{Elliot Creager}, \bibinfo{person}{David
  Madras}, \bibinfo{person}{Toniann Pitassi}, {and} \bibinfo{person}{Richard
  Zemel}.} \bibinfo{year}{2020}\natexlab{}.
\newblock \showarticletitle{Causal modeling for fairness in dynamical systems}.
  In \bibinfo{booktitle}{\emph{International Conference on Machine Learning}}.
  PMLR, \bibinfo{pages}{2185--2195}.
\newblock


\bibitem[D'Amour et~al\mbox{.}(2020)]%
        {d2020fairness}
\bibfield{author}{\bibinfo{person}{Alexander D'Amour}, \bibinfo{person}{Hansa
  Srinivasan}, \bibinfo{person}{James Atwood}, \bibinfo{person}{Pallavi
  Baljekar}, \bibinfo{person}{David Sculley}, {and} \bibinfo{person}{Yoni
  Halpern}.} \bibinfo{year}{2020}\natexlab{}.
\newblock \showarticletitle{Fairness is not static: deeper understanding of
  long term fairness via simulation studies}. In
  \bibinfo{booktitle}{\emph{Proceedings of the 2020 Conference on Fairness,
  Accountability, and Transparency}}. \bibinfo{pages}{525--534}.
\newblock


\bibitem[Dwork et~al\mbox{.}(2012)]%
        {dwork2012fairness}
\bibfield{author}{\bibinfo{person}{Cynthia Dwork}, \bibinfo{person}{Moritz
  Hardt}, \bibinfo{person}{Toniann Pitassi}, \bibinfo{person}{Omer Reingold},
  {and} \bibinfo{person}{Richard Zemel}.} \bibinfo{year}{2012}\natexlab{}.
\newblock \showarticletitle{Fairness through awareness}. In
  \bibinfo{booktitle}{\emph{Proceedings of the 3rd innovations in theoretical
  computer science conference}}. \bibinfo{pages}{214--226}.
\newblock


\bibitem[Elmalaki(2021)]%
        {elmalaki2021fair}
\bibfield{author}{\bibinfo{person}{Salma Elmalaki}.}
  \bibinfo{year}{2021}\natexlab{}.
\newblock \showarticletitle{Fair-iot: Fairness-aware human-in-the-loop
  reinforcement learning for harnessing human variability in personalized iot}.
  In \bibinfo{booktitle}{\emph{Proceedings of the International Conference on
  Internet-of-Things Design and Implementation}}. \bibinfo{pages}{119--132}.
\newblock


\bibitem[Elmalaki(2022)]%
        {elmalaki2022maconauto}
\bibfield{author}{\bibinfo{person}{Salma Elmalaki}.}
  \bibinfo{year}{2022}\natexlab{}.
\newblock \showarticletitle{MAConAuto: Framework for Mobile-Assisted
  Human-in-the-Loop Automotive System}. In \bibinfo{booktitle}{\emph{2022 IEEE
  Intelligent Vehicles Symposium (IV)}}. IEEE, \bibinfo{pages}{740--749}.
\newblock


\bibitem[Elmalaki et~al\mbox{.}(2018a)]%
        {elmalaki2018internet}
\bibfield{author}{\bibinfo{person}{Salma Elmalaki}, \bibinfo{person}{Yasser
  Shoukry}, {and} \bibinfo{person}{Mani Srivastava}.}
  \bibinfo{year}{2018}\natexlab{a}.
\newblock \showarticletitle{Internet of personalized and autonomous things
  (IOPAT) smart homes case study}. In \bibinfo{booktitle}{\emph{Proceedings of
  the 1st ACM International Workshop on Smart Cities and Fog Computing}}.
  \bibinfo{pages}{35--40}.
\newblock


\bibitem[Elmalaki et~al\mbox{.}(2018b)]%
        {elmalaki2018sentio}
\bibfield{author}{\bibinfo{person}{Salma Elmalaki}, \bibinfo{person}{Huey-Ru
  Tsai}, {and} \bibinfo{person}{Mani Srivastava}.}
  \bibinfo{year}{2018}\natexlab{b}.
\newblock \showarticletitle{Sentio: Driver-in-the-loop forward collision
  warning using multisample reinforcement learning}. In
  \bibinfo{booktitle}{\emph{Proceedings of the 16th ACM Conference on Embedded
  Networked Sensor Systems}}. \bibinfo{pages}{28--40}.
\newblock


\bibitem[Fanger(1970)]%
        {fanger1970thermal}
\bibfield{author}{\bibinfo{person}{Poul~O Fanger}.}
  \bibinfo{year}{1970}\natexlab{}.
\newblock \showarticletitle{Thermal comfort. Analysis and applications in
  environmental engineering.}
\newblock \bibinfo{journal}{\emph{Thermal comfort. Analysis and applications in
  environmental engineering.}} (\bibinfo{year}{1970}).
\newblock


\bibitem[Friedler et~al\mbox{.}(2019)]%
        {friedler2019comparative}
\bibfield{author}{\bibinfo{person}{Sorelle~A Friedler}, \bibinfo{person}{Carlos
  Scheidegger}, \bibinfo{person}{Suresh Venkatasubramanian},
  \bibinfo{person}{Sonam Choudhary}, \bibinfo{person}{Evan~P Hamilton}, {and}
  \bibinfo{person}{Derek Roth}.} \bibinfo{year}{2019}\natexlab{}.
\newblock \showarticletitle{A comparative study of fairness-enhancing
  interventions in machine learning}. In \bibinfo{booktitle}{\emph{Proceedings
  of the conference on fairness, accountability, and transparency}}.
  \bibinfo{pages}{329--338}.
\newblock


\bibitem[Gerber(2014)]%
        {gerber2014energyplus}
\bibfield{author}{\bibinfo{person}{Michael Gerber}.}
  \bibinfo{year}{2014}\natexlab{}.
\newblock \showarticletitle{energyplus energy Simulation Software}.
\newblock  (\bibinfo{year}{2014}).
\newblock


\bibitem[Gillen et~al\mbox{.}(2018)]%
        {gillen2018online}
\bibfield{author}{\bibinfo{person}{Stephen Gillen},
  \bibinfo{person}{Christopher Jung}, \bibinfo{person}{Michael Kearns}, {and}
  \bibinfo{person}{Aaron Roth}.} \bibinfo{year}{2018}\natexlab{}.
\newblock \showarticletitle{Online learning with an unknown fairness metric}.
  In \bibinfo{booktitle}{\emph{Advances in neural information processing
  systems}}. \bibinfo{pages}{2600--2609}.
\newblock


\bibitem[Goel et~al\mbox{.}(2018)]%
        {goel2018non}
\bibfield{author}{\bibinfo{person}{Naman Goel}, \bibinfo{person}{Mohammad
  Yaghini}, {and} \bibinfo{person}{Boi Faltings}.}
  \bibinfo{year}{2018}\natexlab{}.
\newblock \showarticletitle{Non-discriminatory machine learning through convex
  fairness criteria}. In \bibinfo{booktitle}{\emph{Proceedings of the 2018
  AAAI/ACM Conference on AI, Ethics, and Society}}. \bibinfo{pages}{116--116}.
\newblock


\bibitem[Hadfield-Menell et~al\mbox{.}(2016)]%
        {hadfield2016cooperative}
\bibfield{author}{\bibinfo{person}{Dylan Hadfield-Menell},
  \bibinfo{person}{Stuart~J Russell}, \bibinfo{person}{Pieter Abbeel}, {and}
  \bibinfo{person}{Anca Dragan}.} \bibinfo{year}{2016}\natexlab{}.
\newblock \showarticletitle{Cooperative inverse reinforcement learning}. In
  \bibinfo{booktitle}{\emph{Advances in neural information processing
  systems}}. \bibinfo{pages}{3909--3917}.
\newblock


\bibitem[Hardt et~al\mbox{.}(2016)]%
        {hardt2016equality}
\bibfield{author}{\bibinfo{person}{Moritz Hardt}, \bibinfo{person}{Eric Price},
  {and} \bibinfo{person}{Nati Srebro}.} \bibinfo{year}{2016}\natexlab{}.
\newblock \showarticletitle{Equality of opportunity in supervised learning}.
\newblock \bibinfo{journal}{\emph{Advances in neural information processing
  systems}}  \bibinfo{volume}{29} (\bibinfo{year}{2016}).
\newblock


\bibitem[Hashimoto et~al\mbox{.}(2018)]%
        {hashimoto2018fairness}
\bibfield{author}{\bibinfo{person}{Tatsunori~B Hashimoto},
  \bibinfo{person}{Megha Srivastava}, \bibinfo{person}{Hongseok Namkoong},
  {and} \bibinfo{person}{Percy Liang}.} \bibinfo{year}{2018}\natexlab{}.
\newblock \showarticletitle{Fairness without demographics in repeated loss
  minimization}.
\newblock \bibinfo{journal}{\emph{arXiv preprint arXiv:1806.08010}}
  (\bibinfo{year}{2018}).
\newblock


\bibitem[Hoffman et~al\mbox{.}(2018)]%
        {hoffman2018discretion}
\bibfield{author}{\bibinfo{person}{Mitchell Hoffman}, \bibinfo{person}{Lisa~B
  Kahn}, {and} \bibinfo{person}{Danielle Li}.} \bibinfo{year}{2018}\natexlab{}.
\newblock \showarticletitle{Discretion in hiring}.
\newblock \bibinfo{journal}{\emph{The Quarterly Journal of Economics}}
  \bibinfo{volume}{133}, \bibinfo{number}{2} (\bibinfo{year}{2018}),
  \bibinfo{pages}{765--800}.
\newblock


\bibitem[Hughes et~al\mbox{.}(2018)]%
        {hughes2018inequity}
\bibfield{author}{\bibinfo{person}{Edward Hughes}, \bibinfo{person}{Joel~Z
  Leibo}, \bibinfo{person}{Matthew Phillips}, \bibinfo{person}{Karl Tuyls},
  \bibinfo{person}{Edgar Due{\~n}ez-Guzman},
  \bibinfo{person}{Antonio~Garc{\'\i}a Casta{\~n}eda}, \bibinfo{person}{Iain
  Dunning}, \bibinfo{person}{Tina Zhu}, \bibinfo{person}{Kevin McKee},
  \bibinfo{person}{Raphael Koster}, {et~al\mbox{.}}}
  \bibinfo{year}{2018}\natexlab{}.
\newblock \showarticletitle{Inequity aversion improves cooperation in
  intertemporal social dilemmas}. In \bibinfo{booktitle}{\emph{Advances in
  neural information processing systems}}. \bibinfo{pages}{3326--3336}.
\newblock


\bibitem[Ib{\'a}{\~n}ez et~al\mbox{.}(2014)]%
        {ibanez2014experimenting}
\bibfield{author}{\bibinfo{person}{Mar{\'\i}a~Blanca Ib{\'a}{\~n}ez},
  \bibinfo{person}{{\'A}ngela Di~Serio}, \bibinfo{person}{Diego Villar{\'a}n},
  {and} \bibinfo{person}{Carlos~Delgado Kloos}.}
  \bibinfo{year}{2014}\natexlab{}.
\newblock \showarticletitle{Experimenting with electromagnetism using augmented
  reality: Impact on flow student experience and educational effectiveness}.
\newblock \bibinfo{journal}{\emph{Computers \& Education}}
  \bibinfo{volume}{71} (\bibinfo{year}{2014}), \bibinfo{pages}{1--13}.
\newblock


\bibitem[Iglesias and Blanco(2008)]%
        {iglesias2008new}
\bibfield{author}{\bibinfo{person}{Eva Iglesias} {and}
  \bibinfo{person}{Mar{\'\i}a Blanco}.} \bibinfo{year}{2008}\natexlab{}.
\newblock \showarticletitle{New directions in water resources management: The
  role of water pricing policies}.
\newblock \bibinfo{journal}{\emph{Water Resources Research}}
  \bibinfo{volume}{44}, \bibinfo{number}{6} (\bibinfo{year}{2008}).
\newblock


\bibitem[Irizarry et~al\mbox{.}(2013)]%
        {irizarry2013infospot}
\bibfield{author}{\bibinfo{person}{Javier Irizarry}, \bibinfo{person}{Masoud
  Gheisari}, \bibinfo{person}{Graceline Williams}, {and}
  \bibinfo{person}{Bruce~N Walker}.} \bibinfo{year}{2013}\natexlab{}.
\newblock \showarticletitle{InfoSPOT: A mobile Augmented Reality method for
  accessing building information through a situation awareness approach}.
\newblock \bibinfo{journal}{\emph{Automation in construction}}
  \bibinfo{volume}{33} (\bibinfo{year}{2013}), \bibinfo{pages}{11--23}.
\newblock


\bibitem[Jabbari et~al\mbox{.}(2017)]%
        {jabbari2017fairness}
\bibfield{author}{\bibinfo{person}{Shahin Jabbari}, \bibinfo{person}{Matthew
  Joseph}, \bibinfo{person}{Michael Kearns}, \bibinfo{person}{Jamie
  Morgenstern}, {and} \bibinfo{person}{Aaron Roth}.}
  \bibinfo{year}{2017}\natexlab{}.
\newblock \showarticletitle{Fairness in reinforcement learning}. In
  \bibinfo{booktitle}{\emph{International Conference on Machine Learning}}.
  \bibinfo{pages}{1617--1626}.
\newblock


\bibitem[Jackson(2005)]%
        {jackson2005motivating}
\bibfield{author}{\bibinfo{person}{Tim Jackson}.}
  \bibinfo{year}{2005}\natexlab{}.
\newblock \showarticletitle{Motivating sustainable consumption}.
\newblock \bibinfo{journal}{\emph{Sustainable Development Research Network}}
  \bibinfo{volume}{29}, \bibinfo{number}{1} (\bibinfo{year}{2005}),
  \bibinfo{pages}{30--40}.
\newblock


\bibitem[Jiang and Lu(2019)]%
        {jiang2019learning}
\bibfield{author}{\bibinfo{person}{Jiechuan Jiang} {and}
  \bibinfo{person}{Zongqing Lu}.} \bibinfo{year}{2019}\natexlab{}.
\newblock \showarticletitle{Learning fairness in multi-agent systems}. In
  \bibinfo{booktitle}{\emph{Advances in Neural Information Processing
  Systems}}. \bibinfo{pages}{13854--13865}.
\newblock


\bibitem[Jorgensen et~al\mbox{.}(2009)]%
        {jorgensen2009household}
\bibfield{author}{\bibinfo{person}{Bradley Jorgensen},
  \bibinfo{person}{Michelle Graymore}, {and} \bibinfo{person}{Kevin O'Toole}.}
  \bibinfo{year}{2009}\natexlab{}.
\newblock \showarticletitle{Household water use behavior: An integrated model}.
\newblock \bibinfo{journal}{\emph{Journal of environmental management}}
  \bibinfo{volume}{91}, \bibinfo{number}{1} (\bibinfo{year}{2009}),
  \bibinfo{pages}{227--236}.
\newblock


\bibitem[Joseph et~al\mbox{.}(2016)]%
        {joseph2016fairness}
\bibfield{author}{\bibinfo{person}{Matthew Joseph}, \bibinfo{person}{Michael
  Kearns}, \bibinfo{person}{Jamie~H Morgenstern}, {and} \bibinfo{person}{Aaron
  Roth}.} \bibinfo{year}{2016}\natexlab{}.
\newblock \showarticletitle{Fairness in learning: Classic and contextual
  bandits}. In \bibinfo{booktitle}{\emph{Advances in Neural Information
  Processing Systems}}. \bibinfo{pages}{325--333}.
\newblock


\bibitem[Jung and Jazizadeh(2017)]%
        {jung2017towards}
\bibfield{author}{\bibinfo{person}{Wooyoung Jung} {and}
  \bibinfo{person}{Farrokh Jazizadeh}.} \bibinfo{year}{2017}\natexlab{}.
\newblock \showarticletitle{Towards integration of doppler radar sensors into
  personalized thermoregulation-based control of HVAC}. In
  \bibinfo{booktitle}{\emph{Proceedings of the 4th ACM International Conference
  on Systems for Energy-Efficient Built Environments}}. ACM,
  \bibinfo{pages}{21}.
\newblock


\bibitem[Kannan et~al\mbox{.}(2018)]%
        {kannan2018smoothed}
\bibfield{author}{\bibinfo{person}{Sampath Kannan}, \bibinfo{person}{Jamie~H
  Morgenstern}, \bibinfo{person}{Aaron Roth}, \bibinfo{person}{Bo Waggoner},
  {and} \bibinfo{person}{Zhiwei~Steven Wu}.} \bibinfo{year}{2018}\natexlab{}.
\newblock \showarticletitle{A smoothed analysis of the greedy algorithm for the
  linear contextual bandit problem}. In \bibinfo{booktitle}{\emph{Advances in
  Neural Information Processing Systems}}. \bibinfo{pages}{2227--2236}.
\newblock


\bibitem[Kannan et~al\mbox{.}(2019)]%
        {kannan2019downstream}
\bibfield{author}{\bibinfo{person}{Sampath Kannan}, \bibinfo{person}{Aaron
  Roth}, {and} \bibinfo{person}{Juba Ziani}.} \bibinfo{year}{2019}\natexlab{}.
\newblock \showarticletitle{Downstream effects of affirmative action}. In
  \bibinfo{booktitle}{\emph{Proceedings of the Conference on Fairness,
  Accountability, and Transparency}}. \bibinfo{pages}{240--248}.
\newblock


\bibitem[Khargonekar and Sampath(2020)]%
        {khargonekar2020framework}
\bibfield{author}{\bibinfo{person}{Pramod~P Khargonekar} {and}
  \bibinfo{person}{Meera Sampath}.} \bibinfo{year}{2020}\natexlab{}.
\newblock \showarticletitle{A framework for ethics in cyber-physical-human
  systems}.
\newblock \bibinfo{journal}{\emph{IFAC-PapersOnLine}} \bibinfo{volume}{53},
  \bibinfo{number}{2} (\bibinfo{year}{2020}), \bibinfo{pages}{17008--17015}.
\newblock


\bibitem[Kleinberg et~al\mbox{.}(2018)]%
        {kleinberg2018algorithmic}
\bibfield{author}{\bibinfo{person}{Jon Kleinberg}, \bibinfo{person}{Jens
  Ludwig}, \bibinfo{person}{Sendhil Mullainathan}, {and}
  \bibinfo{person}{Ashesh Rambachan}.} \bibinfo{year}{2018}\natexlab{}.
\newblock \showarticletitle{Algorithmic fairness}. In
  \bibinfo{booktitle}{\emph{Aea papers and proceedings}},
  Vol.~\bibinfo{volume}{108}. \bibinfo{pages}{22--27}.
\newblock


\bibitem[Kusner et~al\mbox{.}(2017)]%
        {kusner2017counterfactual}
\bibfield{author}{\bibinfo{person}{Matt~J Kusner}, \bibinfo{person}{Joshua
  Loftus}, \bibinfo{person}{Chris Russell}, {and} \bibinfo{person}{Ricardo
  Silva}.} \bibinfo{year}{2017}\natexlab{}.
\newblock \showarticletitle{Counterfactual fairness}.
\newblock \bibinfo{journal}{\emph{Advances in neural information processing
  systems}}  \bibinfo{volume}{30} (\bibinfo{year}{2017}).
\newblock


\bibitem[Lin(1991)]%
        {lin1991divergence}
\bibfield{author}{\bibinfo{person}{Jianhua Lin}.}
  \bibinfo{year}{1991}\natexlab{}.
\newblock \showarticletitle{Divergence measures based on the Shannon entropy}.
\newblock \bibinfo{journal}{\emph{IEEE Transactions on Information theory}}
  \bibinfo{volume}{37}, \bibinfo{number}{1} (\bibinfo{year}{1991}),
  \bibinfo{pages}{145--151}.
\newblock


\bibitem[Liu et~al\mbox{.}(2018)]%
        {liu2018delayed}
\bibfield{author}{\bibinfo{person}{Lydia~T Liu}, \bibinfo{person}{Sarah Dean},
  \bibinfo{person}{Esther Rolf}, \bibinfo{person}{Max Simchowitz}, {and}
  \bibinfo{person}{Moritz Hardt}.} \bibinfo{year}{2018}\natexlab{}.
\newblock \showarticletitle{Delayed impact of fair machine learning}. In
  \bibinfo{booktitle}{\emph{International Conference on Machine Learning}}.
  PMLR, \bibinfo{pages}{3150--3158}.
\newblock


\bibitem[MATLAB(2023)]%
        {MATLABwater}
\bibfield{author}{\bibinfo{person}{MATLAB}.} \bibinfo{year}{2023}\natexlab{}.
\newblock \bibinfo{title}{Water Distribution System Scheduling Using
  Reinforcement Learning}.
\newblock
\newblock
\urldef\tempurl%
\url{https://www.mathworks.com/help/reinforcement-learning/ug/water-distribution-scheduling-system.html}
\showURL{%
Retrieved June 15, 2023 from \tempurl}


\bibitem[Mehrabi et~al\mbox{.}(2021)]%
        {mehrabi2021survey}
\bibfield{author}{\bibinfo{person}{Ninareh Mehrabi}, \bibinfo{person}{Fred
  Morstatter}, \bibinfo{person}{Nripsuta Saxena}, \bibinfo{person}{Kristina
  Lerman}, {and} \bibinfo{person}{Aram Galstyan}.}
  \bibinfo{year}{2021}\natexlab{}.
\newblock \showarticletitle{A survey on bias and fairness in machine learning}.
\newblock \bibinfo{journal}{\emph{ACM Computing Surveys (CSUR)}}
  \bibinfo{volume}{54}, \bibinfo{number}{6} (\bibinfo{year}{2021}),
  \bibinfo{pages}{1--35}.
\newblock


\bibitem[Milli et~al\mbox{.}(2019)]%
        {milli2019social}
\bibfield{author}{\bibinfo{person}{Smitha Milli}, \bibinfo{person}{John
  Miller}, \bibinfo{person}{Anca~D Dragan}, {and} \bibinfo{person}{Moritz
  Hardt}.} \bibinfo{year}{2019}\natexlab{}.
\newblock \showarticletitle{The social cost of strategic classification}. In
  \bibinfo{booktitle}{\emph{Proceedings of the Conference on Fairness,
  Accountability, and Transparency}}. \bibinfo{pages}{230--239}.
\newblock


\bibitem[Mnih et~al\mbox{.}(2013)]%
        {mnih2013playing}
\bibfield{author}{\bibinfo{person}{Volodymyr Mnih}, \bibinfo{person}{Koray
  Kavukcuoglu}, \bibinfo{person}{David Silver}, \bibinfo{person}{Alex Graves},
  \bibinfo{person}{Ioannis Antonoglou}, \bibinfo{person}{Daan Wierstra}, {and}
  \bibinfo{person}{Martin Riedmiller}.} \bibinfo{year}{2013}\natexlab{}.
\newblock \showarticletitle{Playing atari with deep reinforcement learning}.
\newblock \bibinfo{journal}{\emph{arXiv preprint arXiv:1312.5602}}
  (\bibinfo{year}{2013}).
\newblock


\bibitem[Mukerjee et~al\mbox{.}(2002)]%
        {mukerjee2002multi}
\bibfield{author}{\bibinfo{person}{Amitabha Mukerjee}, \bibinfo{person}{Rita
  Biswas}, \bibinfo{person}{Kalyanmoy Deb}, {and} \bibinfo{person}{Amrit~P
  Mathur}.} \bibinfo{year}{2002}\natexlab{}.
\newblock \showarticletitle{Multi--objective evolutionary algorithms for the
  risk--return trade--off in bank loan management}.
\newblock \bibinfo{journal}{\emph{International Transactions in operational
  research}} \bibinfo{volume}{9}, \bibinfo{number}{5} (\bibinfo{year}{2002}),
  \bibinfo{pages}{583--597}.
\newblock


\bibitem[Northpointe(2015)]%
        {northpointe2015practitioner}
\bibfield{author}{\bibinfo{person}{Northpointe}.}
  \bibinfo{year}{2015}\natexlab{}.
\newblock \bibinfo{title}{Practitioner’s Guide to {COMPAS} Core}.
\newblock
  \bibinfo{howpublished}{\url{https://s3.documentcloud.org/documents/2840784/Practitioner-s-Guide-to-COMPAS-Core.pdf}}.
\newblock


\bibitem[{Philadelphia Government}(2023)]%
        {water1mojtaba}
\bibfield{author}{\bibinfo{person}{{Philadelphia Government}}.}
  \bibinfo{year}{2023}\natexlab{}.
\newblock \bibinfo{title}{{``Gallons Used Per Person Per Day''}}.
\newblock
  \bibinfo{howpublished}{\url{https://water.phila.gov/pool/files/home-water-use-ig5.pdf}}.
\newblock


\bibitem[Picard(2000)]%
        {picard2000affective}
\bibfield{author}{\bibinfo{person}{Rosalind~W Picard}.}
  \bibinfo{year}{2000}\natexlab{}.
\newblock \bibinfo{booktitle}{\emph{Affective computing}}.
\newblock \bibinfo{publisher}{MIT press}.
\newblock


\bibitem[Rambachan et~al\mbox{.}(2020)]%
        {rambachan2020economic}
\bibfield{author}{\bibinfo{person}{Ashesh Rambachan}, \bibinfo{person}{Jon
  Kleinberg}, \bibinfo{person}{Jens Ludwig}, {and} \bibinfo{person}{Sendhil
  Mullainathan}.} \bibinfo{year}{2020}\natexlab{}.
\newblock \showarticletitle{An economic perspective on algorithmic fairness}.
  In \bibinfo{booktitle}{\emph{AEA Papers and Proceedings}},
  Vol.~\bibinfo{volume}{110}. \bibinfo{pages}{91--95}.
\newblock


\bibitem[Ratliff et~al\mbox{.}(2018)]%
        {ratliff2018perspective}
\bibfield{author}{\bibinfo{person}{Lillian~J Ratliff}, \bibinfo{person}{Roy
  Dong}, \bibinfo{person}{Shreyas Sekar}, {and} \bibinfo{person}{Tanner Fiez}.}
  \bibinfo{year}{2018}\natexlab{}.
\newblock \showarticletitle{A perspective on incentive design: Challenges and
  opportunities}.
\newblock \bibinfo{journal}{\emph{Annual Review of Control, Robotics, and
  Autonomous Systems}} \bibinfo{volume}{2}, \bibinfo{number}{1}
  (\bibinfo{year}{2018}), \bibinfo{pages}{1--34}.
\newblock


\bibitem[Ratliff and Fiez(2020)]%
        {ratliff2020adaptive}
\bibfield{author}{\bibinfo{person}{Lillian~J Ratliff} {and}
  \bibinfo{person}{Tanner Fiez}.} \bibinfo{year}{2020}\natexlab{}.
\newblock \showarticletitle{Adaptive incentive design}.
\newblock \bibinfo{journal}{\emph{IEEE Trans. Automat. Control}}
  \bibinfo{volume}{66}, \bibinfo{number}{8} (\bibinfo{year}{2020}),
  \bibinfo{pages}{3871--3878}.
\newblock


\bibitem[Reddy et~al\mbox{.}(2017)]%
        {reddy2017accelerating}
\bibfield{author}{\bibinfo{person}{Siddharth Reddy}, \bibinfo{person}{Sergey
  Levine}, {and} \bibinfo{person}{Anca Dragan}.}
  \bibinfo{year}{2017}\natexlab{}.
\newblock \showarticletitle{Accelerating human learning with deep reinforcement
  learning}. In \bibinfo{booktitle}{\emph{NIPS workshop: teaching machines,
  robots, and humans}}.
\newblock


\bibitem[Sadigh et~al\mbox{.}(2017)]%
        {sadigh2017active}
\bibfield{author}{\bibinfo{person}{Dorsa Sadigh}, \bibinfo{person}{Anca~D
  Dragan}, \bibinfo{person}{Shankar Sastry}, {and} \bibinfo{person}{Sanjit~A
  Seshia}.} \bibinfo{year}{2017}\natexlab{}.
\newblock \showarticletitle{Active Preference-Based Learning of Reward
  Functions.}. In \bibinfo{booktitle}{\emph{Robotics: Science and Systems}}.
\newblock


\bibitem[Sechi et~al\mbox{.}(2013)]%
        {sechi2013water}
\bibfield{author}{\bibinfo{person}{Giovanni~M Sechi}, \bibinfo{person}{Riccardo
  Zucca}, {and} \bibinfo{person}{Paola Zuddas}.}
  \bibinfo{year}{2013}\natexlab{}.
\newblock \showarticletitle{Water costs allocation in complex systems using a
  cooperative game theory approach}.
\newblock \bibinfo{journal}{\emph{Water Resources Management}}
  \bibinfo{volume}{27} (\bibinfo{year}{2013}), \bibinfo{pages}{1781--1796}.
\newblock


\bibitem[Shin et~al\mbox{.}(2017)]%
        {shin2017exploring}
\bibfield{author}{\bibinfo{person}{Eun-Jeong Shin}, \bibinfo{person}{Roberto
  Yus}, \bibinfo{person}{Sharad Mehrotra}, {and} \bibinfo{person}{Nalini
  Venkatasubramanian}.} \bibinfo{year}{2017}\natexlab{}.
\newblock \showarticletitle{Exploring fairness in participatory thermal comfort
  control in smart buildings}. In \bibinfo{booktitle}{\emph{Proceedings of the
  4th ACM International Conference on Systems for Energy-Efficient Built
  Environments}}. \bibinfo{pages}{1--10}.
\newblock


\bibitem[Siddique et~al\mbox{.}(2020)]%
        {siddique2020learning}
\bibfield{author}{\bibinfo{person}{Umer Siddique}, \bibinfo{person}{Paul Weng},
  {and} \bibinfo{person}{Matthieu Zimmer}.} \bibinfo{year}{2020}\natexlab{}.
\newblock \showarticletitle{Learning fair policies in multi-objective (deep)
  reinforcement learning with average and discounted rewards}. In
  \bibinfo{booktitle}{\emph{International Conference on Machine Learning}}.
  PMLR, \bibinfo{pages}{8905--8915}.
\newblock


\bibitem[Sutton and Barto(2018)]%
        {sutton2018reinforcement}
\bibfield{author}{\bibinfo{person}{Richard~S Sutton} {and}
  \bibinfo{person}{Andrew~G Barto}.} \bibinfo{year}{2018}\natexlab{}.
\newblock \bibinfo{booktitle}{\emph{Reinforcement learning: An introduction}}.
\newblock \bibinfo{publisher}{MIT press}.
\newblock


\bibitem[Sutton et~al\mbox{.}(1999)]%
        {sutton1999between}
\bibfield{author}{\bibinfo{person}{Richard~S Sutton}, \bibinfo{person}{Doina
  Precup}, {and} \bibinfo{person}{Satinder Singh}.}
  \bibinfo{year}{1999}\natexlab{}.
\newblock \showarticletitle{Between MDPs and semi-MDPs: A framework for
  temporal abstraction in reinforcement learning}.
\newblock \bibinfo{journal}{\emph{Artificial intelligence}}
  \bibinfo{volume}{112}, \bibinfo{number}{1-2} (\bibinfo{year}{1999}),
  \bibinfo{pages}{181--211}.
\newblock


\bibitem[Sztipanovits et~al\mbox{.}(2019)]%
        {sztipanovits2019science}
\bibfield{author}{\bibinfo{person}{Janos Sztipanovits},
  \bibinfo{person}{Xenofon Koutsoukos}, \bibinfo{person}{Gabor Karsai},
  \bibinfo{person}{Shankar Sastry}, \bibinfo{person}{Claire Tomlin},
  \bibinfo{person}{Werner Damm}, \bibinfo{person}{Martin Fr{\"a}nzle},
  \bibinfo{person}{Jochem Rieger}, \bibinfo{person}{Alexander Pretschner},
  {and} \bibinfo{person}{Frank K{\"o}ster}.} \bibinfo{year}{2019}\natexlab{}.
\newblock \showarticletitle{Science of design for societal-scale cyber-physical
  systems: challenges and opportunities}.
\newblock \bibinfo{journal}{\emph{Cyber-Physical Systems}} \bibinfo{volume}{5},
  \bibinfo{number}{3} (\bibinfo{year}{2019}), \bibinfo{pages}{145--172}.
\newblock


\bibitem[Taherisadr et~al\mbox{.}(2023)]%
        {taherisadr2023adaparl}
\bibfield{author}{\bibinfo{person}{Mojtaba Taherisadr},
  \bibinfo{person}{Stelios~Andrew Stavroulakis}, {and} \bibinfo{person}{Salma
  Elmalaki}.} \bibinfo{year}{2023}\natexlab{}.
\newblock \showarticletitle{adaPARL: Adaptive Privacy-Aware Reinforcement
  Learning for Sequential-Decision Making Human-in-the-Loop Systems}.
\newblock \bibinfo{journal}{\emph{arXiv preprint arXiv:2303.04257}}
  (\bibinfo{year}{2023}).
\newblock


\bibitem[Terai et~al\mbox{.}(2020)]%
        {terai2020detecting}
\bibfield{author}{\bibinfo{person}{Shogo Terai}, \bibinfo{person}{Shizuka
  Shirai}, \bibinfo{person}{Mehrasa Alizadeh}, \bibinfo{person}{Ryosuke
  Kawamura}, \bibinfo{person}{Noriko Takemura}, \bibinfo{person}{Yuki
  Uranishi}, \bibinfo{person}{Haruo Takemura}, {and} \bibinfo{person}{Hajime
  Nagahara}.} \bibinfo{year}{2020}\natexlab{}.
\newblock \showarticletitle{Detecting learner drowsiness based on facial
  expressions and head movements in online courses}. In
  \bibinfo{booktitle}{\emph{Proceedings of the 25th International Conference on
  Intelligent User Interfaces Companion}}. \bibinfo{pages}{124--125}.
\newblock


\bibitem[Yu et~al\mbox{.}(2019)]%
        {yu2019deep}
\bibfield{author}{\bibinfo{person}{Yiding Yu}, \bibinfo{person}{Taotao Wang},
  {and} \bibinfo{person}{Soung~Chang Liew}.} \bibinfo{year}{2019}\natexlab{}.
\newblock \showarticletitle{Deep-reinforcement learning multiple access for
  heterogeneous wireless networks}.
\newblock \bibinfo{journal}{\emph{IEEE Journal on Selected Areas in
  Communications}} \bibinfo{volume}{37}, \bibinfo{number}{6}
  (\bibinfo{year}{2019}), \bibinfo{pages}{1277--1290}.
\newblock


\end{thebibliography}
